\documentclass[]{relaxed_system_lab}

\usepackage[utf8]{inputenc}
\usepackage[T1]{fontenc}
\usepackage{geometry}
\usepackage{amsmath,amssymb}  %
\usepackage{charter}

\usepackage[toc,page,header]{appendix}

\usepackage{minitoc}
\usepackage{cleveref} 
\usepackage{subcaption}
\usepackage{booktabs}
\usepackage{graphicx}
\usepackage{pgfplots}
\usepackage{pgfplotstable}
\usepackage{xcolor}
\usepackage{CJKutf8}
\usetikzlibrary{patterns}
\usepackage{float} 
\usepackage{caption}
\usepackage{bm}

\usepackage{soul} %
\usepackage{algpseudocode}  %
\usepackage{parskip}

\definecolor{cvprblue}{rgb}{0.21,0.49,0.74}
\definecolor{fallbackgreen}{rgb}{130, 180, 102}
\definecolor{stopred}{rgb}{251, 225, 224}

\ifdefined\final
\usepackage[disable]{todonotes}
\else
\usepackage[textsize=tiny]{todonotes}
\fi

\usepackage{natbib}
\usepackage{latexsym}

\usepackage{url}
\usepackage{amssymb}
\usepackage[utf8]{inputenc}
\usepackage{microtype}
\usepackage{booktabs}
\usepackage{pifont} 
\usepackage{multirow}
\usepackage{makecell}
\usepackage{paralist}
\usepackage{xspace}
\usepackage{color}
\usepackage{xcolor}
\usepackage{colortbl}
\usepackage{adjustbox}
\usepackage{hyperref} 
\usepackage[edges]{forest}
\usepackage{tikz} 
\usepackage{caption}
\usepackage{amsfonts}

\hypersetup{
    colorlinks,
    linkcolor={blue!80!black},
    citecolor={blue!80!black},
}
\tikzset{
    root/.style =             {align=center, text width=1cm, rounded corners=3pt, line width=0.3mm, fill=gray!10, draw=gray!80, font=\small},
    demographic/.style =         {align=center, text width=1.8cm, rounded corners=3pt, line width=0.3mm, fill=blue!10, draw=blue!80, font=\footnotesize},
    demographic_work/.style =    {align=center, text width=10cm, rounded corners=3pt, line width=0.3mm, fill=blue!10, draw=blue!0, font=\footnotesize},
    character/.style =         {align=center, text width=1.8cm, rounded corners=3pt, line width=0.3mm, fill=red!10, draw=red!80, font=\footnotesize},
    character_work/.style =    {align=center, text width=10cm, rounded corners=3pt, line width=0.3mm, fill=red!10, draw=red!0, font=\footnotesize},
    personalization/.style =           {align=center, text width=1.8cm, rounded corners=3pt, line width=0.3mm, fill=cyan!10, draw=cyan!80, font=\footnotesize},
    personalization_work/.style =      {align=center, text width=10cm, rounded corners=3pt, line width=0.3mm, fill=cyan!10, draw=cyan!0, font=\footnotesize},
    risk/.style =         {align=center, text width=1.8cm, rounded corners=3pt, line width=0.3mm, fill=orange!10, draw=orange!80, font=\footnotesize},
    risk_work/.style =    {align=center, text width=10cm, rounded corners=3pt, line width=0.3mm, fill=orange!10, draw=orange!0, font=\footnotesize},
}

\usepackage{CJK}

\newtcolorbox{promptbox}[1][]{
  enhanced,
  breakable,
  colback=promptboxlightgray,
  colframe=promptboxblue!30,
  arc=8pt,
  boxrule=0.5pt,
  left=12pt,
  right=12pt,
  top=8pt,
  bottom=8pt,
  fonttitle=\bfseries,
  fontupper=\linespread{1.2}\selectfont,
  title=#1
}

\usepackage{hyperref}
\usepackage{url}
\usepackage[linesnumbered,ruled,vlined]{algorithm2e}
\usepackage[most]{tcolorbox}
\usepackage{booktabs}
\usepackage{caption} 
\usepackage{graphicx}
\usepackage{array}
\usepackage{enumitem}
\usepackage{xspace}
\usepackage{xcolor}

\definecolor{wbPurpleBorder}{HTML}{5C449C}
\definecolor{wbPurpleBack}{HTML}{F5E7FC}
\definecolor{wbPurpleMid}{HTML}{A586D1}

\newcommand{\name}{\textsc{S2SServiceBench}\xspace}
\newcommand{\nproducts}{10\xspace}
\newcommand{\ncases}{150+\xspace}
\newcommand{\ntasks}{500\xspace}

\title{\name: A Multimodal Benchmark for Last-Mile S2S Climate Services}

\author{Chenyue Li$^{1*}$, Wen Deng$^{1*}$, Zhuotao Sun$^1$, Mengxi Jin$^1$,  Hanzhe Cui$^1$, Han Li$^2$, Shentong Li$^3$, Man Kit Yu$^1$, Ming Long Lai$^1$, Yuhao Yang, Mengqian Lu$^1$, Binhang Yuan$^{1\dagger}$}

\affiliation{$^1$The Hong Kong University of Science and Technology \\ $^2$Nanjing University of Information Science and Technology $^3$Beijing Normal University \\ \small $^*$Equal contribution, $^{\dagger}$Corresponding author}

\abstract{Subseasonal-to-seasonal (S2S) forecasts play an essential role in providing a decision-critical weeks-to-months planning window for climate resilience and sustainability, yet a growing bottleneck is the last-mile gap: translating scientific forecasts into trusted, actionable climate services, requiring reliable multimodal understanding and decision-facing reasoning under uncertainty. 
Meanwhile, multimodal large language models (MLLMs) and corresponding agentic paradigms have made rapid progress in supporting various workflows, but it remains unclear whether they can reliably generate decision-making deliverables from operational service products (e.g., actionable signal comprehension, decision-making handoff, and decision analysis \& planning) under uncertainty. 
We introduce \name, a multimodal benchmark for last-mile S2S climate services curated from an operational climate-service system to evaluate this capability. 
\name covers 10 service products with about \ncases expert-selected cases in total, spanning six application domains---Agriculture, Disasters, Energy, Finance, Health, and Shipping.
Each case is instantiated at three service levels, yielding around \ntasks tasks and 1{,}000+ evaluation items across climate resilience and sustainability applications. 
Using \name, we benchmark state-of-the-art MLLMs and agents, and analyze performance across products and service levels, revealing persistent challenges in S2S service plot understanding and reasoning---namely, actionable signal comprehension, operationalizing uncertainty into executable handoffs, and stable, evidence-grounded analysis and planning for dynamic hazards-while offering actionable guidance for building future climate-service agents.}

\begin{document}

\maketitle

%%%%%%%%%%%%%%%%%%%%%%%%%%%%%%%%%%%%%%%%%%%%%%%%%%%%%%%%%%%%%%%%%%%%%%%%%%%%%%%%%%%%%%%%%%%%%%%%
%%%%%%%%%%%%%%%%%%%%%%%%%%%%%%%%%%%%%%%%%%%%%%%%%%%%%%%%%%%%%%%%%%%%%%%%%%%%%%%%%%%%%%%%%%%%%%%%

\section{Introduction}

\begin{figure}[t]
    \centering
    \includegraphics[width=\linewidth]{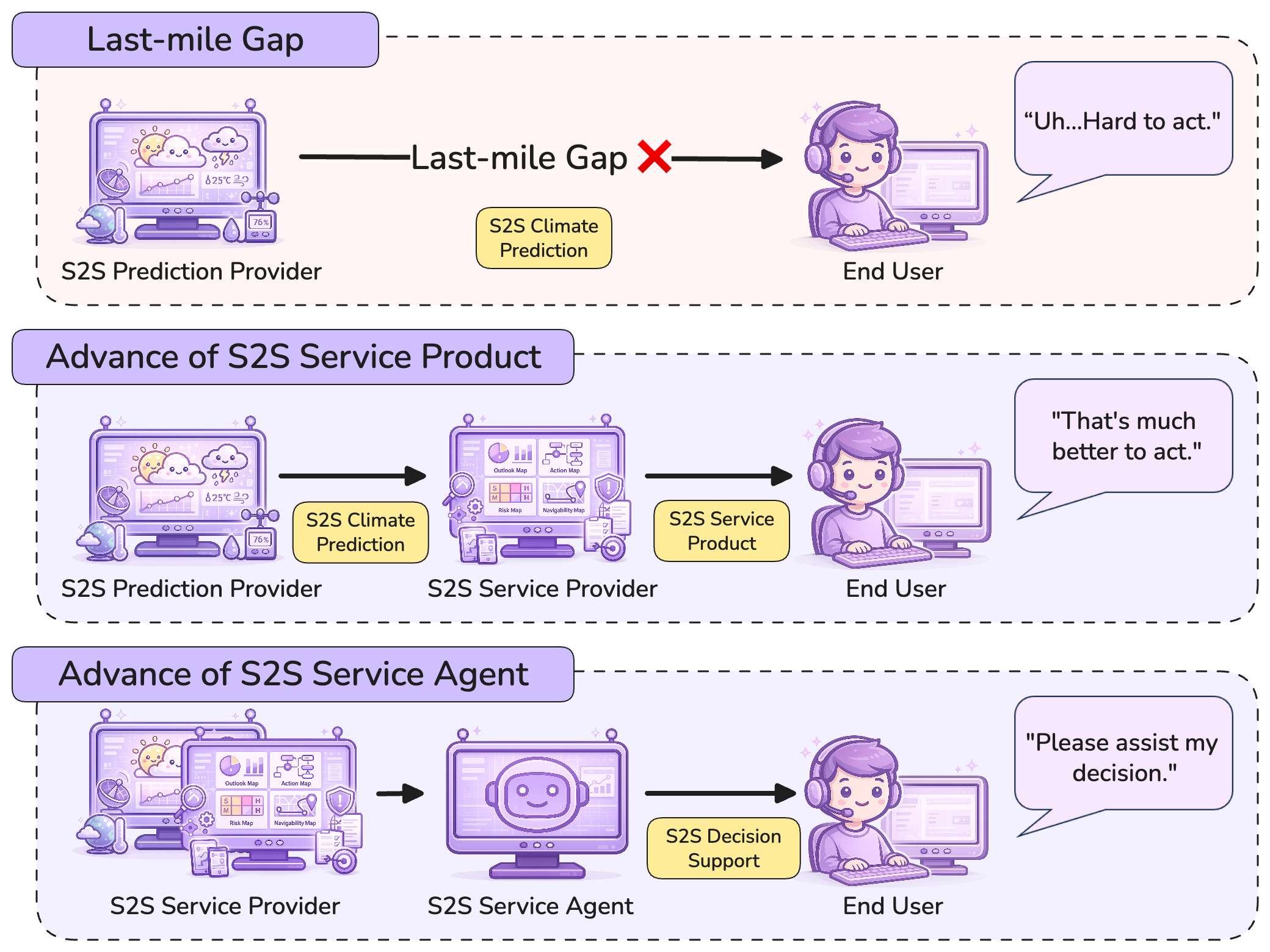}
    \caption{The overview of Subseasonal-to-seasonal (S2S) scope. Top panel: the ``last-mile'' gap---S2S prediction providers struggle to tailor predictions and communicate confidence, while users struggle to interpret uncertainty and act on it~\cite{yang2026last}. Middle panel: this gap is partially bridged by curating S2S predictions into S2S service products, which better support users' decisions. Bottom panel: to further narrow the gap, an S2S service agent can interact with these service products and deliver actionable S2S decision support for end users.}
    \label{fig:s2s_overview}
    \vspace{-0.5em}
\end{figure}

Subseasonal-to-seasonal (S2S) prediction has evolved to be equipped with an operational capability over the past decade, and the community focus is increasingly shifting from producing better climate prediction to delivering better S2S services, with the goal of narrowing the ``last-mile'' gap~\cite{yang2026last} to provide better support for the emerging ecosystem of service-oriented products. To bridge this gap, we expect that a service agent is needed to \textbf{(\underline{i})} interpret these service products and \textbf{(\underline{ii})} translate them into actionable S2S decision support that can effectively assist end users (Figure~\ref{fig:s2s_overview}). \footnote{Additional background is provided in Appendix~\ref{sec:appendix_forecast_to_products}.}

We believe that the state-of-the-art multimodal LLMs and the corresponding agentic deployment create a great opportunity: an S2S Service Agent could operate on top of existing service products (Fig.~\ref{fig:s2s_overview}) to produce decision support deliverables (e.g., briefings, alerts, and constraint-aware recommended actions for disaster adaptation/mitigation and resource management). 
Crucially, such deliverables require three core service capabilities implied by operational S2S products:
\textbf{(\underline{i})} \textit{actionable signal comprehension}---extracting time-localized, decision-relevant signals in a MLLM (agents) consumable form;
\textbf{(\underline{ii})} \textit{decision-making handoff}---turning product evidence into executable response guidance with clear triggers, constraints, and uncertainty-conditioned scenario branching; and
\textbf{(\underline{iii})} \textit{decision analysis \& planning}---producing evidence-grounded strategic interpretation of spatiotemporal patterns and planning-oriented considerations without over-claiming beyond the product.
Motivated by this need, we ask a prerequisite question:

% \vspace{-0.75em} 
\begin{quote}
    \textit{Can current MLLMs or corresponding agents reliably deliver these three core capabilities across products and service levels when operating on top of operational service products?}
\end{quote}
% \vspace{-0.75em} 

% \textbf{\name.}
To address this, we introduce \name, a multimodal benchmark for last-mile S2S climate services curated from newly emerging service products from an operational climate-service system. Curating from an operational climate-service system ensures that tasks reflect real product designs, confidence conventions, and workflow constraints that determine whether the resulting handoffs are actionable for end users.
\name covers 10 service products with about \ncases expert-selected cases in total, spanning six application domains---Agriculture, Disasters, Energy, Finance, Health, and Shipping.
\footnote{A single operational product can support multiple decision contexts; we therefore allow products to be associated with more than one domain label when applicable.}
Each case is instantiated at three service levels, yielding roughly \ntasks tasks and 1{,}000+ evaluation items across climate resilience and sustainability applications. 
Tasks are grounded in realistic multimodal service products (e.g., action maps, outlook plots), reflecting how service providers actually interact with forecasting information in practice.
The three service levels are designed to progressively test capabilities from product reading, to interpretation with uncertainty, to decision-facing deliverables that mimic operational handoffs to end users.

In summary, our contributions are listed as follows:

\textbf{\underline{Contribution 1.}} 
    % \item \textbf{\name: an operationally grounded benchmark for decision support deliverables.}
We curate \name from an operational climate-service system, covering 10 realistic multimodal S2S production services across six domains (i.e., Agriculture, Disasters, Energy, Finance, Health, Shipping).
Grounded in practitioner-facing artifacts (outlook/action/risk maps) and expert-selected operational cases, \name enables rigorous evaluation of whether models can generate decision support deliverables from existing service products under uncertainty.

\textbf{\underline{Contribution 2.}} 
% \textbf{A three-capability, three-level formulation with fine-grained evaluation.}
We operationalize decision support deliverables via three core service capabilities---actionable signal comprehension, decision-making handoff, and decision analysis \& planning---and instantiate each case at three corresponding service task levels.
This yields roughly \ntasks tasks and 1{,}000+ evaluation items with unified schema-constrained deliverables, enabling diagnosis across products and capability levels.

\textbf{\underline{Contribution 3.}} 
% \textbf{Empirical diagnosis of current MLLMs/agents and implications for dedicated climate-service agents.}
We benchmark frontier MLLMs under direct prompting and within a standardized agentic workflow, and analyze performance across products, levels, and task formats.
The results reveal strong product-dependent variability and consistent operational bottlenecks (e.g., time localization; trigger/time, feasibility, and uncertainty handling in decision-facing outputs), showing that standardized scaffolds are not a reliable remedy and motivating dedicated climate-service agents with evaluation-aligned guardrails and service-specific tooling.

\begin{figure*}[t]
    \centering
    \begin{subfigure}[t]{0.49\linewidth}
        \centering
        \includegraphics[width=\linewidth]{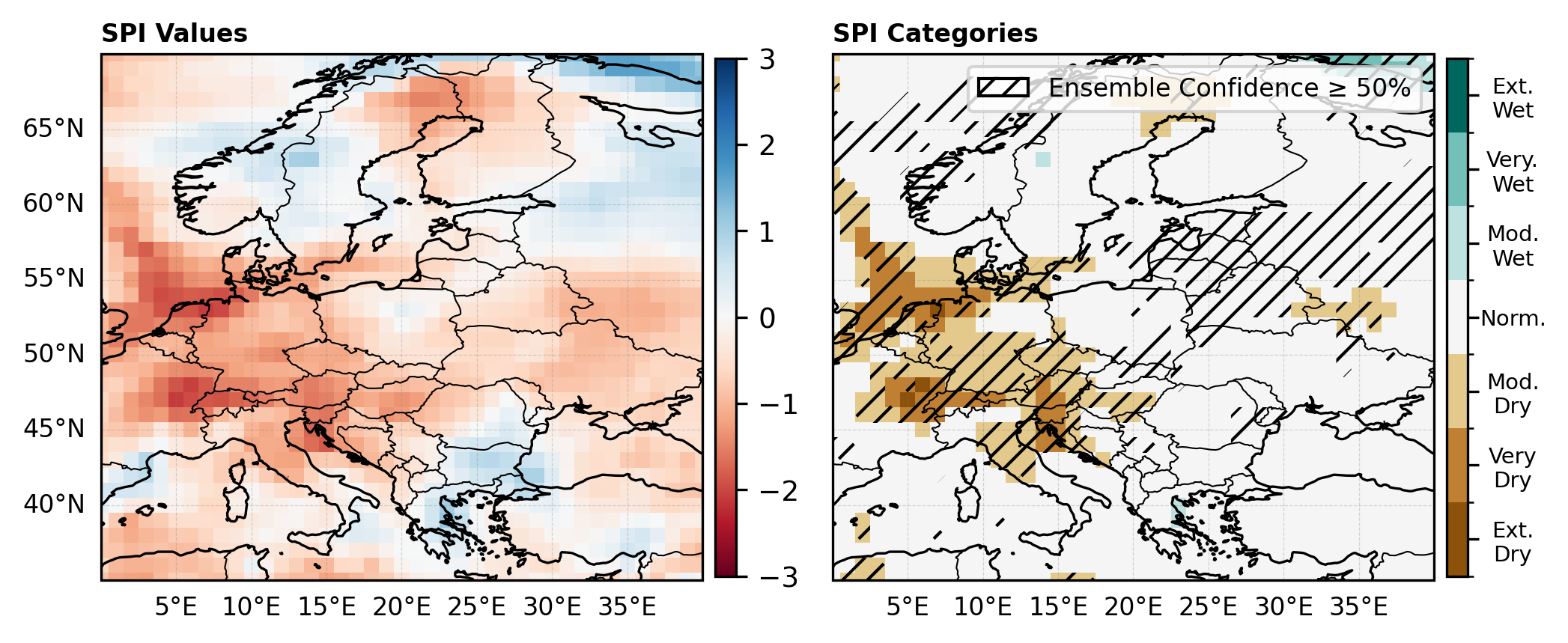}
        % \caption{\textbf{Drought Outlook (SPI-1).} Left: SPI continuous values; right: SPI categories with an ensemble-confidence overlay (hatched; confidence $\ge 0.5$), indicating regions with strong inter-model agreement.}
        \label{fig:spi_example}
    \end{subfigure}
    \hfill
    \begin{subfigure}[t]{0.49\linewidth}
        \centering
        \includegraphics[width=\linewidth]{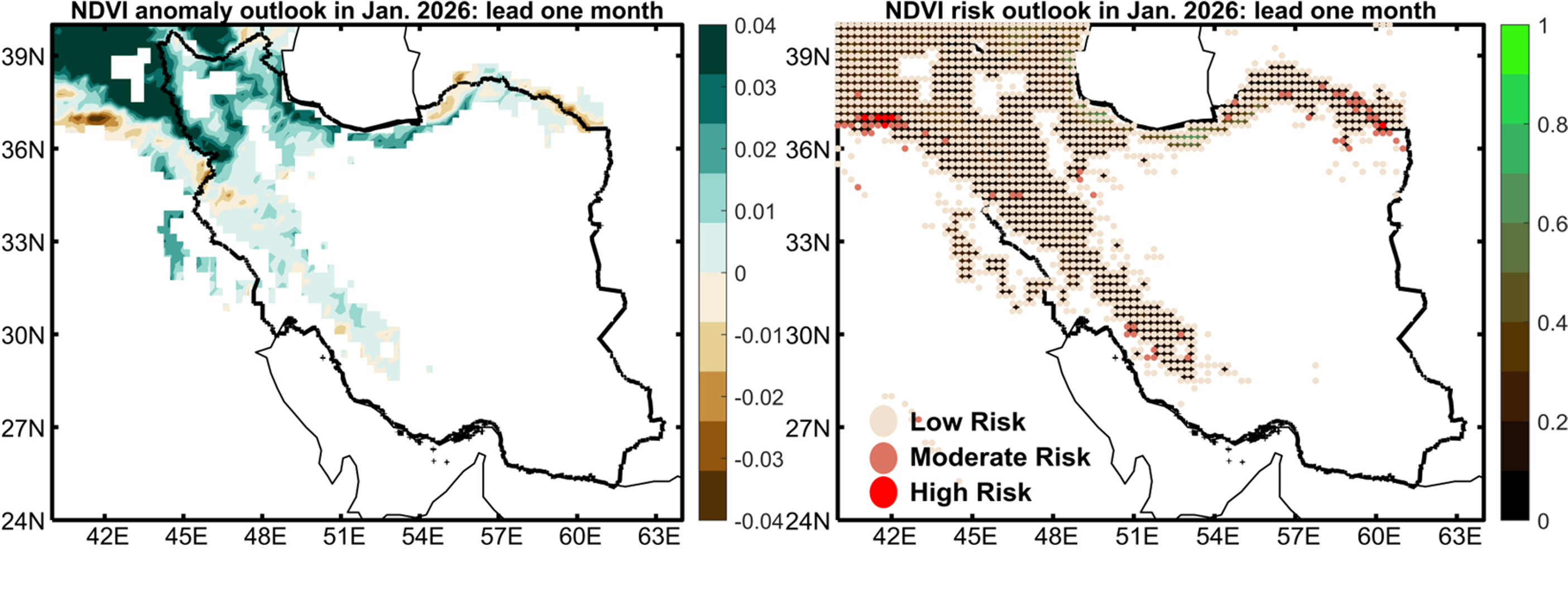}
        % \caption{\textbf{NDVI Outlook.} Left: NDVI anomaly outlook; right: NDVI risk outlook with categorical risk tiers (low/moderate/high) for Jan.\ 2026 at one-month lead.}
        \label{fig:ndvi_example}
    \end{subfigure}
    \vspace{-1em}
    \caption{\textbf{Examples of operational S2S service products used as benchmark inputs.}
    These products illustrate the multimodal structure common in practice: index/anomaly maps paired with categorical tiers and uncertainty cues (e.g., confidence overlays), which models must read and convert into decision-making structured deliverables in \name. Left: \textbf{Drought Outlook (SPI-1).} SPI continuous values and SPI categories with an ensemble-confidence overlay (hatched; confidence $\ge 0.5$), indicating regions with strong inter-model agreement. Right: \textbf{NDVI Outlook.} NDVI anomaly outlook and NDVI risk outlook with categorical risk tiers (low/moderate/high) for Jan.\ 2026 at one-month lead.}
    \label{fig:service_product_examples}
\end{figure*}

%%%%%%%%%%%%%%%%%%%%%%%%%%%%%%%%%%%%%%%%%%%%%%%%%%%%%%%%%%%%%%%%%%%%%%%%%%%%%%%%%%%%%%%%%%%%%%%%
%%%%%%%%%%%%%%%%%%%%%%%%%%%%%%%%%%%%%%%%%%%%%%%%%%%%%%%%%%%%%%%%%%%%%%%%%%%%%%%%%%%%%%%%%%%%%%%%
\vspace{-0.5em}

\section{Related Work}

\textbf{MLLMs and agentic workflows.}
Recent advances in multimodal large language models (MLLMs) have demonstrated steady improvements in multimodal reasoning, enabling multimodal understanding~\cite{openai_gpt52_2025, yue2025mmmu, liu2025medq}, schema-constrained structured generation~\cite{feng2025so, openai_gpt52_2025}, and tool use for retrieval and analysis~\cite{fan2025mcptoolbench++, li2025flow}, with emerging evidence and discussion suggesting potential as assistive tools in related climate and earth science settings~\cite{zhang2025foundation}. 
Building on these foundations, agentic workflows orchestrate planning, retrieval, tool execution, and iterative verification, making them a closer match to real service pipelines than single-turn prompting~\cite{sapkota2025ai, chen2025xbench}. 
Together, these advances make actionable decision support deliverables from operational service products increasingly plausible, yet there remains a lack of service-grounded systematic evaluation that tests reliability under uncertainty and deliverable compliance.

% \textbf{\textcolor{blue}{TBDetermine}agent benchmark}

\textbf{AI for climate science benchmark.}
Recent benchmarks probe LLM/VLM capabilities on weather--climate artifacts and events, including multimodal severe-weather reasoning (WeatherQA)~\cite{ma2024weatherqa}, multimodal event forecasting (CLLMate)~\cite{li2025cllmate}, radar-forecast quality analysis (RadarQA)~\cite{he2025radarqa}, geospatial VLM evaluation (GEOBench-VLM)~\cite{danish2025geobench}, and atmospheric-science reasoning problems (AtmosSci-Bench)~\cite{li2025atmossci}. 
In parallel, S2S benchmarks have primarily focused on improving data-driven prediction and evaluating weather emulators (e.g., ClimaX/ViT, Pangu-Weather, GraphCast, and FourCastNet) under longer lead times and physics-based constraints, as exemplified by ChaosBench~\cite{nathaniel2024chaosbench}. 
Other S2S-oriented benchmarks target observational or reanalysis data quality for model assessment, such as the Nepal multi-source precipitation benchmark that evaluates eleven gridded precipitation products against gauge stations~\cite{nepal2024assessing}. 
While complementary, these benchmarks largely optimize prediction skill or data fidelity, rather than evaluating climate services grounded in operational service products and actionable decision support deliverables under uncertainty---a gap that our benchmark aims to fill.

\textbf{Climate agents.}
Beyond general-purpose agent frameworks, recent work has begun to develop domain-specific agents for Earth and climate science, spanning extreme-weather diagnosis (EWE)~\cite{jiang2025ewe}, end-to-end climate data-science orchestration (CLIMATEAGENT)~\cite{kim2025climateagent}, and cross-modal Earth-observation reasoning with tool ecosystems (Earth-Agent)~\cite{feng2025earth}. 
These systems demonstrate that multi-step planning, tool invocation, and iterative verification can substantially improve scientific analysis workflows, and several efforts introduce dedicated benchmarks to evaluate such agentic capabilities~\cite{jiang2025ewe,kim2025climateagent,feng2025earth}. 
However, they predominantly target automated analysis, discovery, or EO task solving, rather than service-grounded evaluation over operational S2S service products with actionable decision support deliverables and uncertainty communication requirements. 
Our benchmark is complementary to these climate-agent efforts by focusing on last-mile S2S climate services and assessing whether current MLLMs/agents can reliably produce operationally compliant handoffs from real service products.

%%%%%%%%%%%%%%%%%%%%%%%%%%%%%%%%%%%%%%%%%%%%%%%%%%%%%%%%%%%%%%%%%%%%%%%%%%%%%%%%%%%%%%%%%%%%%%%%
%%%%%%%%%%%%%%%%%%%%%%%%%%%%%%%%%%%%%%%%%%%%%%%%%%%%%%%%%%%%%%%%%%%%%%%%%%%%%%%%%%%%%%%%%%%%%%%%

\vspace{-0.5em}
\section{Benchmark Construction}
\label{sec:benchmark_construction}

\begin{figure*}[ht]
    \centering
    \includegraphics[width=1\linewidth]{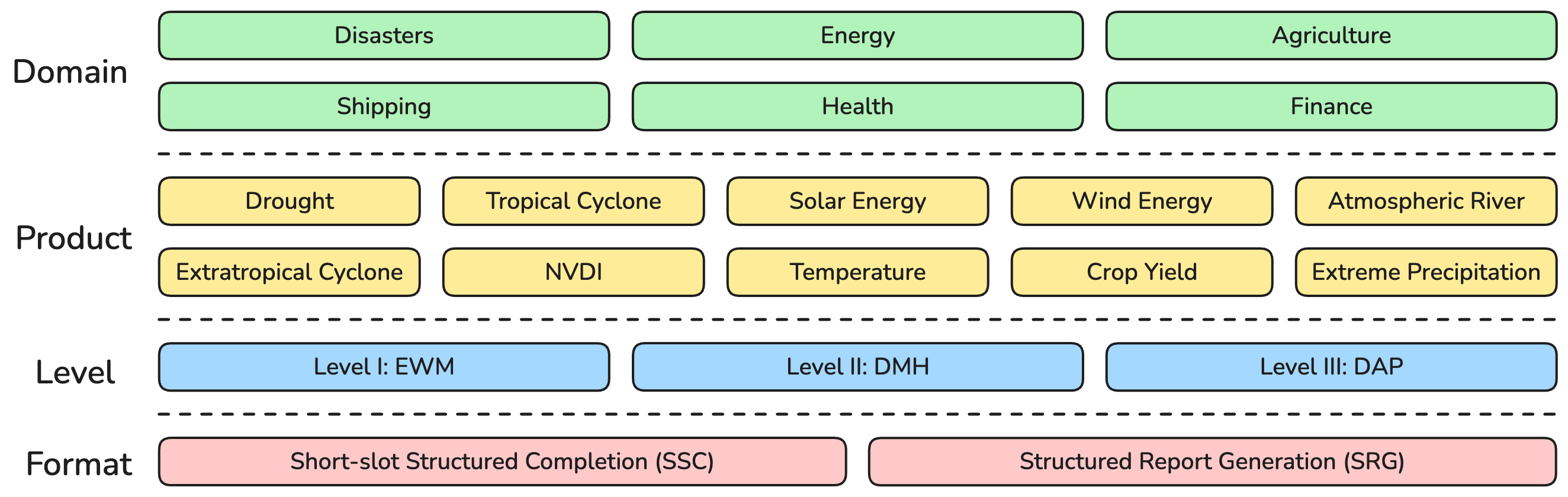}
    \caption{\textbf{Overview of \name.} We curate recurring S2S service products from an operational climate-service system and package each operational instance as a case containing practitioner-facing multimodal artifacts with initialization and valid-time metadata. Each case is instantiated into three service task levels (Section~\ref{sec:taxonomy_and_level}). Tasks are evaluated via (i) SSC with checkable fields and (ii) SRG with schema-constrained report-style outputs graded by a rubric (Appendix~\ref{sec:oed_rubric}), yielding fine-grained evaluation items for diagnosis across products, service levels, and task types.}
    \label{fig:overview}
\end{figure*}

\name is a multimodal benchmark for last-mile S2S climate services, assessing whether current MLLMs and agents can operate on top of operational service products to produce decision support deliverables under uncertainty.
\name is curated from an operational climate-service system: recurring service-product instances are packaged into cases with practitioner-facing multimodal artifacts (outlook/action/risk maps), initialization/valid-time metadata, and expert-verified reference outputs. Figure~\ref{fig:overview} provides a schematic overview of \name, highlighting the benchmark organization from application domains and products to three service levels and two schema-constrained task formats.
Figure~\ref{fig:service_product_examples} shows two representative operational products in \name, illustrating the common service-product structure that drives our evaluation: time-localized signals, impacted regions, categorical risk tiers, and explicit uncertainty cues.

\vspace{-0.5em}
\subsection{Product Taxonomy and Service Task Levels}
\label{sec:taxonomy_and_level}

\textbf{From operational products to benchmark tasks.}
We \textbf{(\underline{i})} identify recurring product templates in the operational system, \textbf{(\underline{ii})} select representative operational instances as benchmark cases with multimodal artifacts and metadata, and \textbf{(\underline{iii})} instantiate each case into three service task levels to probe progressively more decision-facing behavior.

\textbf{Product taxonomy.}
\name includes \textit{10 recurring operational service products} spanning six high-impact application domains---\textit{Agriculture, Disasters, Energy, Finance, Health, and Shipping}.
In practice, a single service product often supports multiple decision contexts; we therefore allow products to be associated with more than one domain label when appropriate.
The full product catalog and domain coverage are provided in Appendix~\ref{sec:appendix_product_catalog}.

\textbf{Service levels.}
To evaluate whether models can generate decision support deliverables from operational service products, we instantiate each case at three service task levels that target distinct capabilities:
\textbf{Level I}: \textit{Signal Comprehension --- Early Warning \& Mitigation (EWM)},
\textbf{Level II}: \textit{Action Handoff --- Decision-making Handoff (DMH)}, and
\textbf{Level III}: \textit{Strategic Assessment --- Decision Analysis \& Planning (DAP)}.
Full level definitions are provided in Appendix~\ref{sec:appendix_levels}.

% \section{Design Principles}
% \label{sec:appendix_design_principles}
\textbf{Design principles.}
\name is built around four design principles that directly operationalize the requirements of last-mile S2S services discussed in Section~1:
\textbf{(\underline{i}) Service-grounded realism}: tasks are derived from operational products and conventions, preserving real visualization templates, confidence qualifiers, and workflow constraints.
\textbf{(\underline{ii}) Time-localized evaluation}: a substantial portion of tasks require identifying and reporting the correct valid-time window (e.g., week-2--week-4 or month-1), enabling diagnosis of time localization failures that are critical in S2S services.
\textbf{(\underline{iii}) Uncertainty-aware service reasoning}: many tasks require models to represent uncertainty signals from service products (e.g., probabilities, confidence qualifiers, threshold-based risk categories) in structured slots or report fields, rather than treating forecasts as deterministic statements.
\textbf{(\underline{iv}) Deliverable-oriented outputs}: tasks evaluate whether models can produce operationally usable artifacts---including schema-constrained deliverables ranging from short actionable fields to report-style decision handoffs---instead of only answering free-form questions.

\vspace{-0.5em}
\subsection{Benchmark Organization and Scale}
\label{sec:benchmark_organization}

\textbf{From products to cases, tasks, and items.}
Starting from \nproducts recurring operational service products, we curate a set of expert-selected cases, where each case corresponds to a single operational product issuance and includes practitioner-facing multimodal artifacts (outlook/action/risk maps) together with initialization and valid-time metadata.
For each case, we instantiate three level-specific tasks aligned with the three service capabilities needed for decision support deliverables.
Each task is evaluated through schema-constrained structured outputs, which decompose into checkable evaluation items (fields/slots) for fine-grained diagnosis.

\textbf{Benchmark scale.}
Across \nproducts products, we curate around \ncases cases and instantiate each at 3 service levels, yielding \ntasks tasks and 1{,}000+ evaluation items.
A detailed accounting by product and service level is provided in Appendix~\ref{sec:appendix_benchmark_size}.

\subsection{Task Formats and Evaluation}
\label{sec:task_formats}

\textbf{Unified structured deliverables.}
All tasks use schema-constrained structured outputs to mirror operational handoffs.
The key distinction is deliverable complexity: Level~I requires short, machine-consumable slots for reliable signal comprehension, while Levels~II--III use longer decision-facing fields that support guidance and planning under uncertainty.

\textbf{Two schema-constrained task formats.}
We includes two deliverable formats aligned with operational workflows:
\textbf{(1) Short-slot Structured Completion (SSC)} tasks inspired by So-Bench~\cite{feng2025so}, where the model outputs a JSON artifact conforming to a predefined schema with short values (e.g., \texttt{bool}, \texttt{enum}, \texttt{number}, or lists of regions);
and \textbf{(2) Structured Report Generation (SRG)} tasks, where the model outputs a schema-constrained report-style deliverable with predefined fields/slots that can contain longer decision-facing content.
SSC is the primary format for Level~I, while SRG is used in Levels~II--III. Detailed scoring procedures are described in Section~\ref{sec:evaluation}.

%%%%%%%%%%%%%%%%%%%%%%%%%%%%%%%%%%%%%%%%%%%%%%%%%%%%%%%%%%%%%%%%%%%%%%%%%%%%%%%%%%%%%%%%%%%%%%%%
%%%%%%%%%%%%%%%%%%%%%%%%%%%%%%%%%%%%%%%%%%%%%%%%%%%%%%%%%%%%%%%%%%%%%%%%%%%%%%%%%%%%%%%%%%%%%%%%

\section{Evaluation and Experimental Setup}
\label{sec:evaluation}

This section describes \textbf{(\underline{i})} how we score outputs under the two deliverable formats (SSC and SRG), \textbf{(\underline{ii})} the evaluated models, and \textbf{(\underline{iii})} the inference-time protocols (direct prompting vs.\ a standardized agentic workflow) used to benchmark last-mile service readiness.

\subsection{Evaluator Overview}
\label{sec:evaluator_overview}

% \textbf{Two deliverable formats and scoring.}
% Our benchmark uses two schema-constrained deliverable formats (Section~\ref{sec:task_formats}): (i) \textbf{SSC} with short slot values (e.g., \texttt{bool}/\texttt{number}/\texttt{string}) and (ii) \textbf{SRG} with longer decision-facing fields constrained to predefined slots.
% SSC is scored by schema validity plus field-level scoring (with 5\% tolerance for numeric fields), and SRG is graded by a services-oriented rubric (0--5 per dimension) with a minimum structure requirement appended to each SRG prompt.
% Full scoring protocols, field-type handling, coverage-based region matching, and aggregation details are provided in Appendix~\ref{sec:appendix_evaluation_details}.

\paragraph{Two deliverable formats and scoring.}
Our benchmark uses two schema-constrained deliverable formats (Section~\ref{sec:task_formats}): \textbf{(\underline{i})} \textbf{SSC} with short slot values (e.g., \texttt{bool}/\texttt{number}/\texttt{string}) and \textbf{(\underline{ii})} \textbf{SRG} with longer decision-facing fields constrained to predefined slots.
SSC is scored by schema validity plus field-level scoring (with 5\% tolerance for numeric fields).
SRG is graded by a services-oriented rubric (0--5 per dimension) that evaluates:
Context Tailoring (CT; geo/sector/population specificity),
Actionability (ACT; concrete and executable measures),
Trigger \& Time/Horizon Clarity (TTH; operational triggers and timing),
Evidence Grounding (EG; alignment with provided product evidence),
Feasibility \& Constraints (FC; resource/policy/operational constraints and adaptations), and
Uncertainty \& Confidence Handling (UC; confidence-calibrated actions and contingencies).
% We additionally append a minimum structure requirement to each SRG prompt to ensure comparable, scorable outputs across models.
Full scoring protocols, field-type handling, coverage-based region matching, and aggregation details are provided in Appendix~\ref{sec:appendix_evaluation_details}.

\textbf{LLM-as-Judge.}
We adopt LLM-as-Judge for evaluation components that require semantic alignment beyond exact match, including \textbf{(\underline{i})} region/area strings in SSC and \textbf{(\underline{ii})} rubric-based grading for SRG report-style outputs.
Recent work suggests that LLMs can function as effective graders with strong agreement to expert evaluations~\cite{dinh2024sciex}, and several recent benchmarks adopt LLM-as-Judge as a primary evaluation mechanism~\cite{dinh2024sciex,phan2025humanity,feng2025physics}.
To mitigate subjectivity, we use fixed judging protocols and detailed rubrics, following prior practice~\cite{wang2025qmbench}.

\subsection{Models}
\label{sec:models}

% We benchmark six state-of-the-art multimodal LLMs spanning proprietary and open models: \texttt{GPT-5.2} (OpenAI), \texttt{Grok-4} (xAI; \texttt{grok-4-0709}), \texttt{Claude 4.5 Opus} (Anthropic), \texttt{Gemini 3 Pro} (Google), \texttt{Qwen3-VL-32B-Instruct} (Qwen), and \texttt{Llama 4 Maverick Instruct} (Meta).
% Detailed model descriptions and access interfaces are provided in Appendix~\ref{sec:appendix_model_details}.
% \subsection{Model Details}
% \label{sec:appendix_model_details}

We benchmark a set of state-of-the-art multimodal LLMs spanning proprietary and open models:
\begin{itemize}[topsep=2pt,leftmargin=1.2em]
    \item \texttt{GPT-5.2 (OpenAI).} A frontier multimodal model used via the OpenAI API, providing strong general reasoning and robust text--vision understanding for service-facing multimodal tasks.
    \item \texttt{Grok-4 (xAI; grok-4-0709).} A general-purpose multimodal model from xAI, accessed through the xAI API, designed for high-throughput reasoning and multimodal understanding in interactive settings.
    \item \texttt{Claude 4.5 Opus (Anthropic).} Anthropic's flagship Claude model variant emphasizing high-quality reasoning and instruction following, suitable for both direct prompting and agentic scaffolds.
    \item \texttt{Gemini 3 Pro (Google).} A Gemini Pro-family multimodal model accessed via the Gemini API, supporting image-and-text inputs for operational product understanding and decision-oriented generation.
    \item \texttt{Qwen3-VL-32B-Instruct (Qwen).} An open multimodal instruction-tuned vision--language model in the Qwen3-VL family, enabling strong visual grounding and long-form structured generation.
    \item \texttt{Llama 4 Maverick Instruct (Meta).} An open instruction-tuned Llama 4 model (Maverick) intended for strong general reasoning and generation, used as an open-model baseline under the same evaluation protocols.
\end{itemize}

\textbf{Decoding and fairness.}
Unless otherwise specified, we use a unified inference configuration across models: maximum output length of 32{,}768 tokens and default decoding settings for the provider/API.
Within each evaluation setting, all models receive identical task prompts and identical tool availability; only the underlying MLLM is swapped.

\subsection{Evaluation Settings}
\label{sec:evaluation_settings}

We evaluate models under two inference-time protocols: \textbf{direct prompting} and a \textbf{standardized agentic workflow}.
This isolates the effect of lightweight multi-step scaffolding (and tool access) on last-mile S2S service tasks.

\begin{itemize}[topsep=3pt,leftmargin=1.2em]
    \item \textbf{Direct prompting.}
    The MLLM is queried with the task instruction and associated multimodal service artifacts, and must produce the final deliverable in a single response.
    This setting measures standalone capability without external tools or multi-step workflows.

    \item \textbf{Agentic workflow (DeepAgent + web search).}
    We run the same tasks within a standardized Deep Agents scaffold based on LangChain Deep Agents~\cite{langchain_deepagents_overview}.
    The scaffold provides lightweight planning, a file-based workspace for intermediate notes, and tool calls for web retrieval via Tavily search~\cite{tavily_website}.
    % To avoid external leakage and preserve benchmark grounding, we enforce the following constraints:
    % (i) external search is treated as optional background and may not override or contradict the provided service products;
    % (ii) required signals (e.g., valid-time windows, confidence overlays, risk tiers, and impacted regions) must be derived from the given multimodal artifacts;
    % (iii) the final output is scored only for faithfulness to the provided operational products under the same schemas/rubrics as direct prompting (Section~\ref{sec:evaluator_overview} and Appendix~\ref{sec:oed_rubric}).
\end{itemize}

% ------------------------------------------------------------
\subsection{Experimental Questions}
\label{sec:exp_questions}

We design experiments to evaluate whether systems can generate decision support deliverables from operational S2S service products by exercising three core service capabilities---actionable signal comprehension, decision-making handoff, and decision analysis \& planning. Specifically, we ask:

\begin{itemize}[topsep=3pt,leftmargin=1.2em]
    
    \item \textbf{Q1 (Capability-level performance).}
    How well do current MLLMs generate decision support deliverables across the three capability levels (signal comprehension, decision-making handoff, and decision analysis \& planning) and across service products?

    \item \textbf{Q2 (Operational bottlenecks in decision support deliverables).}
    For decision-facing deliverables, which rubric dimensions are most challenging (e.g., trigger \& time/horizon clarity, feasibility/constraints, uncertainty handling, evidence grounding), and how do these bottlenecks vary across products and capability levels?

    \item \textbf{Q3 (Direct prompting vs.\ agentic workflows).}
    Does a standardized agentic workflow improve decision support deliverables over direct prompting, and for which products and capability levels does it help or hurt?

\end{itemize}

%%%%%%%%%%%%%%%%%%%%%%%%%%%%%%%%%%%%%%%%%%%%%%%%%%%%%%%%%%%%%%%%%%%%%%%%%%%%%%%%%%%%%%%%%%%%%%%%
%%%%%%%%%%%%%%%%%%%%%%%%%%%%%%%%%%%%%%%%%%%%%%%%%%%%%%%%%%%%%%%%%%%%%%%%%%%%%%%%%%%%%%%%%%%%%%%%

\section{Results and Analysis}

\begin{table*}[!t]
\caption{\textbf{Direct prompting performance across service products and service levels.}
Abbreviations: AR = Atmospheric River; Drgt = Drought; ETC = Extratropical Cyclone; EPrp = Extreme Precipitation; NDVI = Normalized Difference Vegetation Index; Sol = Solar Energy; TC = Tropical Cyclone; Temp = Temperature; Wind = Wind Energy; CYld = Crop Yield; Ovr = Overall.}
\label{tab:level_model_topics_direct}
\centering
\large
\begin{adjustbox}{max width=\textwidth}
\begin{tabular}{l l c c c c c c c c c c | c}
\toprule
\textbf{Row Group} & \textbf{Model}
& \textbf{Drgt} & \textbf{TC} & \textbf{Sol} & \textbf{Wind} & \textbf{ETC} & \textbf{NDVI}
& \textbf{Temp} & \textbf{CYld} & \textbf{EPrp} & \textbf{AR} & \textbf{Ovr} \\
\midrule
\multirow{6}{*}{\textbf{Level 1}} 
& GPT-5.2 & 0.1578 & 0.1393 & 0.2416 & 0.3193 & 0.3375 & 0.2947 & 0.4938 & 0.2809 & 0.8308 & 0.5175 & 0.3613 \\
& Grok-4 & 0.0722 & 0.0660 & 0.1070 & 0.1367 & 0.3125 & 0.1077 & 0.0875 & 0.1208 & 0.5385 & 0.4474 & 0.1996 \\
& Claude 4.5 Opus & 0.1177 & 0.1467 & 0.1544 & 0.2238 & 0.3375 & 0.1958 & 0.3937 & 0.1895 & 0.2369 & 0.7368 & 0.2733 \\
& Gemini 3 Pro & 0.1517 & 0.1287 & 0.2154 & 0.2208 & 0.3625 & 0.3298 & 0.5250 & 0.2431 & 0.7692 & 0.6574 & 0.3604 \\
& Qwen3-VL-32B-Instruct & 0.1057 & 0.0653 & 0.1105 & 0.2596 & 0.3750 & 0.2220 & 0.3187 & 0.1918 & 0.5385 & 0.4035 & 0.2591 \\
& Llama 4 Maverick Instruct & 0.0911 & 0.0747 & 0.1235 & 0.2677 & 0.3875 & 0.2438 & 0.3312 & 0.1813 & 0.4615 & 0.4474 & 0.2610 \\
\midrule
\multirow{6}{*}{\textbf{Level 2}} 
& GPT-5.2 & 0.8275 & 0.7640 & 0.7643 & 0.5938 & 0.6350 & 0.2917 & 0.9150 & 0.5574 & 0.3595 & 0.6708 & 0.6379 \\
& Grok-4 & 0.4772 & 0.4240 & 0.4119 & 0.4313 & 0.4483 & 0.3433 & 0.5400 & 0.2213 & 0.2571 & 0.1170 & 0.3671 \\
& Claude 4.5 Opus & 0.6423 & 0.6307 & 0.5631 & 0.4000 & 0.4850 & 0.5267 & 0.8417 & 0.3917 & 0.2857 & 0.2871 & 0.5054 \\
& Gemini 3 Pro & 0.5090 & 0.5107 & 0.4786 & 0.4083 & 0.3467 & 0.2617 & 0.5930 & 0.3241 & 0.2881 & 0.0716 & 0.3792 \\
& Qwen3-VL-32B-Instruct & 0.4122 & 0.4653 & 0.4655 & 0.3479 & 0.2733 & 0.3200 & 0.6583 & 0.2065 & 0.2714 & 0.1012 & 0.3522 \\
& Llama 4 Maverick Instruct & 0.2413 & 0.2133 & 0.2226 & 0.2063 & 0.1567 & 0.2133 & 0.3433 & 0.1917 & 0.1429 & 0.0825 & 0.2014 \\
\midrule
\multirow{6}{*}{\textbf{Level 3}} 
& GPT-5.2 & 0.8008 & 0.2640 & 0.5548 & 0.5708 & 0.1733 & 0.3489 & 0.4800 & 0.7648 & 0.8119 & 0.3184 & 0.5088 \\
& Grok-4 & 0.3492 & 0.1933 & 0.3857 & 0.4000 & 0.1233 & 0.4400 & 0.3883 & 0.2954 & 0.4167 & 0.1424 & 0.3134 \\
& Claude 4.5 Opus & 0.6016 & 0.2187 & 0.4869 & 0.3938 & 0.1600 & 0.5983 & 0.4650 & 0.4806 & 0.5405 & 0.2360 & 0.4181 \\
& Gemini 3 Pro & 0.4508 & 0.2067 & 0.3571 & 0.2708 & 0.1367 & 0.3817 & 0.4517 & 0.3296 & 0.4262 & 0.2566 & 0.3268 \\
& Qwen3-VL-32B-Instruct & 0.3151 & 0.1806 & 0.3905 & 0.3417 & 0.1567 & 0.4467 & 0.4550 & 0.2185 & 0.4357 & 0.2015 & 0.3142 \\
& Llama 4 Maverick Instruct & 0.2317 & 0.0467 & 0.2762 & 0.2146 & 0.0900 & 0.2850 & 0.2783 & 0.2176 & 0.2333 & 0.1427 & 0.2016 \\
\bottomrule
\end{tabular}
\end{adjustbox}
\end{table*}

% ==================== Q1 ====================
\subsection{MLLM Capability Evaluation Results}
\label{sec:results_q1}

Table~\ref{tab:level_model_topics_direct} reports end-to-end results across 10 operational service products and three service task levels, testing whether current MLLMs can generate decision support deliverables via three core capabilities: actionable signal comprehension, decision-making handoff, and decision analysis \& planning.
Overall, performance is strongly level- and product-dependent: the same model can perform well on one product/capability but degrade substantially on another, indicating that last-mile service capability cannot be captured by a single ``multimodal reasoning'' score.

\textbf{Overall ranking across capability levels.}
At \textbf{Level~I} (SSC signal comprehension), GPT-5.2 and Gemini 3 Pro achieve the strongest overall scores (Ovr $\approx$ 0.36), yet absolute performance remains low.
At \textbf{Level~II} (SRG decision-making handoff), GPT-5.2 is best overall (Ovr $\approx$ 0.64), followed by Claude 4.5 Opus (Ovr $\approx$ 0.51).
At \textbf{Level~III} (SRG decision analysis \& planning), GPT-5.2 remains strongest (Ovr $\approx$ 0.51), with markedly larger topic-to-topic variance.

\textbf{Signal comprehension remains difficult despite short outputs.}
Although Level~I requires only short-slot structured completion, overall accuracy remains low even for top models, suggesting that the bottleneck is service plot understanding, checkable extraction from operational products.
% The wide spread across products at Level~I (e.g., Drgt/TC vs.\ EPrp/AR) further indicates that small differences in product presentation and encoding can materially affect extraction reliability.

\textbf{Product heterogeneity across decision-facing capabilities.}
Scores vary widely across products at Levels~II--III, and relative product difficulty can shift across capability levels, indicating that ``easy-to-read'' product cues do not necessarily translate into robust decision support deliverables.
In particular, Level~III exhibits the largest product-driven variance (e.g., strong performance on Drgt/EPrp but much lower on TC/ETC), reflecting persistent difficulty in producing stable, decision-facing planning deliverables for dynamic hazards.

\begin{tcolorbox}[
colback=labbg!7,
colframe=labblue,
  boxrule=0.8pt,
  arc=2mm
]
\textbf{Answer to Question (\underline{i})}:
\textit{Capability-level performance varies strongly by product: even top MLLMs achieve low accuracy on actionable signal comprehension (Level~I), while decision-making handoff (Level~II) and decision analysis \& planning (Level~III) show larger product-driven variance. Overall, reliably generating decision support deliverables from operational service products remains challenging and non-uniform across capabilities.}
\end{tcolorbox}

\begin{table*}[!t]
\caption{Rubric scores for GPT-5.2 at Level 2 and Level 3 (mean$\pm$std, where std=$\sqrt{\text{var}}$). Rubrics: CT = Context Tailoring; ACT = Actionability; TTH = Trigger Time Horizon; EG = Evidence Grounding; FC = Feasibility \& Constraints; UC = Uncertainty \& Confidence.}
\label{tab:level23_rubric_scores_mean_pm_std_gpt52}
\centering
\large
\begin{adjustbox}{max width=\textwidth}
\begin{tabular}{l c c c c c c c c c c c c}
\toprule
\textbf{Topic}
& \multicolumn{6}{c}{\textbf{Level 2 (mean$\pm$std)}}
& \multicolumn{6}{c}{\textbf{Level 3 (mean$\pm$std)}} \\
\cmidrule(lr){2-7}\cmidrule(lr){8-13}
& \textbf{CT} & \textbf{ACT} & \textbf{TTH} & \textbf{EG} & \textbf{FC} & \textbf{UC}
& \textbf{CT} & \textbf{ACT} & \textbf{TTH} & \textbf{EG} & \textbf{FC} & \textbf{UC} \\
\midrule
Atmospheric River
& 4.26$\pm$0.44 & 4.14$\pm$0.44 & 2.53$\pm$0.83 & 3.21$\pm$0.59 & 2.60$\pm$0.56 & 3.39$\pm$0.65
& 2.13$\pm$1.32 & 0.16$\pm$0.37 & 0.39$\pm$0.68 & 3.42$\pm$0.72 & 0.03$\pm$0.16 & 0.42$\pm$0.72 \\
Drought
& 4.40$\pm$0.52 & 4.84$\pm$0.37 & 4.11$\pm$0.65 & 3.70$\pm$0.46 & 3.73$\pm$0.48 & 4.05$\pm$1.08
& 4.10$\pm$0.53 & 4.95$\pm$0.22 & 3.76$\pm$0.69 & 4.07$\pm$0.34 & 3.45$\pm$0.50 & 3.69$\pm$0.56 \\
Extratropical Cyclone
& 4.15$\pm$0.59 & 4.40$\pm$0.50 & 2.45$\pm$1.00 & 4.00$\pm$0.00 & 2.10$\pm$0.45 & 1.95$\pm$0.76
& 1.20$\pm$1.14 & 0.00$\pm$0.00 & 0.00$\pm$0.00 & 4.00$\pm$0.67 & 0.00$\pm$0.00 & 0.00$\pm$0.00 \\
Extreme Precipitation
& 4.21$\pm$0.43 & 0.71$\pm$0.61 & 1.50$\pm$0.65 & 4.00$\pm$0.39 & 0.07$\pm$0.27 & 0.29$\pm$0.47
& 4.64$\pm$0.50 & 5.00$\pm$0.00 & 3.93$\pm$0.62 & 4.14$\pm$0.36 & 3.79$\pm$0.43 & 2.86$\pm$0.77 \\
NDVI
& 4.24$\pm$0.56 & 4.18$\pm$0.81 & 4.12$\pm$0.60 & 4.12$\pm$0.49 & 3.24$\pm$0.66 & 3.71$\pm$0.59
& 4.25$\pm$0.44 & 4.35$\pm$0.49 & 3.30$\pm$0.47 & 4.40$\pm$0.50 & 3.30$\pm$0.57 & 3.30$\pm$0.66 \\
Solar Energy
& 4.64$\pm$0.49 & 4.61$\pm$0.50 & 3.21$\pm$0.83 & 3.93$\pm$0.26 & 3.07$\pm$0.66 & 3.46$\pm$0.69
& 3.57$\pm$0.92 & 2.75$\pm$1.14 & 1.18$\pm$0.61 & 4.04$\pm$0.33 & 2.46$\pm$1.29 & 2.64$\pm$0.87 \\
Wind Energy
& 4.94$\pm$0.25 & 2.88$\pm$0.50 & 0.25$\pm$0.58 & 4.00$\pm$0.00 & 4.00$\pm$0.00 & 1.75$\pm$1.29
& 4.25$\pm$0.58 & 2.81$\pm$0.75 & 0.31$\pm$0.60 & 3.38$\pm$0.50 & 4.06$\pm$0.25 & 2.31$\pm$1.20 \\
Temperature
& 4.40$\pm$0.50 & 5.00$\pm$0.00 & 5.00$\pm$0.00 & 4.85$\pm$0.37 & 3.40$\pm$0.50 & 4.80$\pm$0.41
& 3.30$\pm$0.73 & 1.00$\pm$0.65 & 1.75$\pm$0.55 & 4.70$\pm$0.47 & 0.45$\pm$0.51 & 3.20$\pm$0.41 \\
Tropical Cyclone
& 4.48$\pm$0.59 & 4.96$\pm$0.20 & 3.56$\pm$0.58 & 3.84$\pm$0.55 & 2.96$\pm$0.20 & 3.12$\pm$0.67
& 2.68$\pm$1.11 & 0.00$\pm$0.00 & 0.12$\pm$0.33 & 4.32$\pm$0.48 & 0.00$\pm$0.00 & 0.80$\pm$0.65 \\
Yield
& 3.67$\pm$0.59 & 3.61$\pm$0.55 & 2.00$\pm$0.68 & 2.92$\pm$0.60 & 2.58$\pm$0.97 & 1.94$\pm$1.24
& 4.67$\pm$0.48 & 4.69$\pm$0.47 & 3.36$\pm$0.87 & 3.36$\pm$0.49 & 3.67$\pm$0.53 & 3.19$\pm$0.79 \\
\bottomrule
\end{tabular}
\end{adjustbox}
\end{table*}

% ==================== Q2 ====================
\subsection{Rubric-dimension Bottlenecks}
\label{sec:results_q2_rubric}

To diagnose why generating decision support deliverables remains difficult, we analyze SRG rubric breakdowns for the two decision-facing capabilities: decision-making handoff and decision analysis \& planning.
Tables~\ref{tab:level23_rubric_scores_mean_pm_std_gpt52} report GPT-5.2’s SRG performance by rubric dimension and product topic.

Across both decision-facing levels, Evidence Grounding (EG) is consistently strong (typically $\approx$3.2--4.8), suggesting that GPT-5.2 generally stays aligned with the provided service-product evidence rather than producing ungrounded narratives.
In contrast, the dominant bottlenecks concentrate in operationalization-oriented dimensions that are central to decision support deliverables, especially Trigger \& Time/Horizon clarity (TTH), Feasibility \& Constraints (FC), and Uncertainty \& Confidence handling (UC).
For decision-making handoff, while higher-level framing dimensions such as Context Tailoring (CT) and sometimes Actionability (ACT) can be strong, multiple topics exhibit noticeably lower TTH/FC/UC (e.g., Atmospheric River and Extratropical Cyclone), indicating that deliverable quality is limited less by producing plausible text and more by specifying operational triggers, timing, constraints, and confidence-modulated actions.
For decision analysis \& planning, this gap sharpens: performance becomes highly topic-dependent and can collapse on ACT/TTH/FC/UC for dynamic hazard products such as Tropical Cyclone, Atmospheric River, and Extratropical Cyclone, where these dimensions can approach zero even when EG remains high.
This pattern suggests that the primary difficulty is not multimodal grounding per se, but reliably translating product signals into decision support deliverables with executable triggers, feasible resource-aware actions, and contingency branches under uncertainty.
Finally, the larger standard deviations in several planning-heavy topics (e.g., CT/ACT/UC for AR/TC/Solar/Wind) indicate reduced stability in longer-horizon outputs, reinforcing that robust service planning remains sensitive to consistently meeting operational rubric requirements.

\begin{tcolorbox}[
colback=labbg!7,
colframe=labblue,
boxrule=0.8pt,
arc=2mm
]
\textbf{Answer to Question (\underline{ii})}:
\textit{GPT-5.2 is typically well grounded in the product evidence (high EG), but the main bottlenecks for decision support deliverables are operationalization failures---trigger/time-horizon clarity, feasibility/constraints, and uncertainty/confidence handling---which can collapse for dynamic hazards even when grounding remains strong. These gaps motivate dedicated climate-service agent designs that explicitly enforce operational triggers, constraints, and uncertainty-to-action mapping.}
\end{tcolorbox}

% ==================== Q3 ====================
\subsection{Direct Prompting vs.\ Agentic Paradigm}
\label{sec:results_q3_agent}

Table~\ref{tab:level_topics_gpt52_claude_direct_deepagent} compares direct prompting against a standardized DeepAgent workflow for two representative frontier MLLMs (GPT-5.2 and Claude 4.5 Opus), holding tasks, prompts, and tool availability fixed.
Overall, the standardized agentic scaffold yields non-uniform effects across capabilities: it can modestly help strict signal comprehension, but it does not reliably improve decision support deliverables and can even degrade decision-facing performance.

\textbf{Agent effect by service level (overall).}
At \textbf{Level~I} (signal comprehension via SSC), DeepAgent provides small overall gains for both models (GPT-5.2: Ovr $0.3613\!\rightarrow\!0.3800$; Claude 4.5: $0.2733\!\rightarrow\!0.2990$), consistent with lightweight multi-step scaffolding helping strict slot filling and multimodal product reading.
At \textbf{Level~II} (decision-making handoff via SRG), however, DeepAgent reduces overall performance for both models (GPT-5.2: $0.6379\!\rightarrow\!0.6115$; Claude 4.5: $0.5054\!\rightarrow\!0.4326$), suggesting that generic multi-step workflows can drift from tight evidence/constraint requirements or compound intermediate errors in uncertainty-conditioned action guidance.
At \textbf{Level~III} (decision analysis \& planning via SRG), DeepAgent is also not consistently beneficial: GPT-5.2 improves ($0.5088\!\rightarrow\!0.5588$), but Claude 4.5 substantially degrades ($0.4181\!\rightarrow\!0.3425$), and product-level failures remain pronounced.
Taken together, these results indicate that standardized scaffolding is not a reliable fix for producing decision support deliverables from operational products.

\textbf{Implication: standardized agentic scaffolds have a drawback for decision-facing tasks.}
The mixed and sometimes negative effects at Levels~II--III suggest that generic agent routines (planning + workspace + optional delegation + web search) do not automatically enforce the core requirements of operational climate services under structured output constraints: precise trigger/time specification, calibrated uncertainty-to-action mapping, and faithful grounding to product evidence.
This motivates the need for dedicated climate-service agents with service-specific representations and evaluation-aligned guardrails, rather than relying on standardized general-purpose workflows.

\begin{tcolorbox}[
    colback=labbg!7,
    colframe=labblue,
  boxrule=0.8pt,
  arc=2mm
]
\textbf{Answer to Question (\underline{iii})}:
\textit{A standardized agentic workflow is not a reliable improvement over direct prompting: it yields only small gains for signal comprehension (Level~I), degrades decision-making handoff for both models (Level~II), and is inconsistent for decision analysis \& planning (Level~III). These results highlight a potential drawback of standardized agent scaffolds and suggest that robust decision support deliverables likely require dedicated climate-service agents with tighter evidence- and constraint-aware control.}
\end{tcolorbox}

%%%%%%%%%%%%%%%%%%%%%%%%%%%%%%%%%%%%%%%%%%%%%%%%%%%%%%%%%%%%%%%%%%%%%%%%%%%%%%%%%%%%%%%%%%%%%%%%
%%%%%%%%%%%%%%%%%%%%%%%%%%%%%%%%%%%%%%%%%%%%%%%%%%%%%%%%%%%%%%%%%%%%%%%%%%%%%%%%%%%%%%%%%%%%%%%%

\begin{table*}[!t]
\caption{\textbf{Direct prompting vs.\ standardized agentic workflow for GPT-5.2 and Claude 4.5 Opus across service products and service levels.}
Abbreviations: AR = Atmospheric River; Drgt = Drought; ETC = Extratropical Cyclone; EPrp = Extreme Precipitation; NDVI = Normalized Difference Vegetation Index; Sol = Solar Energy; TC = Tropical Cyclone; Temp = Temperature; Wind = Wind Energy; CYld = Crop Yield; Ovr = Overall.}
\label{tab:level_topics_gpt52_claude_direct_deepagent}
\centering
\large
\begin{adjustbox}{max width=\textwidth}
\begin{tabular}{l l c c c c c c c c c c | c}
\toprule
\textbf{Row Group} & \textbf{Model}
& \textbf{Drgt} & \textbf{TC} & \textbf{Sol} & \textbf{Wind} & \textbf{ETC} & \textbf{NDVI}
& \textbf{Temp} & \textbf{CYld} & \textbf{EPrp} & \textbf{AR} & \textbf{Ovr} \\
\midrule
\multirow{4}{*}{\textbf{Level 1}} 
& GPT-5.2 (Direct)            & 0.1578 & 0.1393 & 0.2416 & 0.3193 & 0.3375 & 0.2947 & 0.4938 & 0.2809 & 0.8308 & 0.5175 & 0.3613 \\
& GPT-5.2 (DeepAgent)         & 0.1832 & 0.1333 & 0.1918 & 0.2784 & 0.3625 & 0.2947 & 0.5500 & 0.2774 & 0.8623 & 0.6667 & 0.3800 \\
& Claude 4.5 Opus (Direct)    & 0.1177 & 0.1467 & 0.1544 & 0.2238 & 0.3375 & 0.1958 & 0.3937 & 0.1895 & 0.2369 & 0.7368 & 0.2733 \\
& Claude 4.5 Opus (DeepAgent) & 0.0926 & 0.1246 & 0.1668 & 0.1818 & 0.3125 & 0.1781 & 0.4313 & 0.2240 & 0.5769 & 0.7018 & 0.2990 \\
\midrule
\multirow{4}{*}{\textbf{Level 2}} 
& GPT-5.2 (Direct)            & 0.8275 & 0.7640 & 0.7643 & 0.5938 & 0.6350 & 0.2917 & 0.9150 & 0.5574 & 0.3595 & 0.6708 & 0.6379 \\
& GPT-5.2 (DeepAgent)         & 0.7481 & 0.7893 & 0.7845 & 0.5896 & 0.6500 & 0.2331 & 0.9117 & 0.5407 & 0.3524 & 0.5158 & 0.6115 \\
& Claude 4.5 Opus (Direct)    & 0.6423 & 0.6307 & 0.5631 & 0.4000 & 0.4850 & 0.5267 & 0.8417 & 0.3917 & 0.2857 & 0.2871 & 0.5054 \\
& Claude 4.5 Opus (DeepAgent) & 0.6810 & 0.2319 & 0.6012 & 0.3479 & 0.4400 & 0.5850 & 0.8283 & 0.3556 & 0.0952 & 0.1602 & 0.4326 \\
\midrule
\multirow{4}{*}{\textbf{Level 3}} 
& GPT-5.2 (Direct)            & 0.8008 & 0.2640 & 0.5548 & 0.5708 & 0.1733 & 0.3489 & 0.4800 & 0.7648 & 0.8119 & 0.3184 & 0.5088 \\
& GPT-5.2 (DeepAgent)         & 0.8000 & 0.2173 & 0.7369 & 0.6271 & 0.1700 & 0.6311 & 0.5050 & 0.7630 & 0.8119 & 0.3257 & 0.5588 \\
& Claude 4.5 Opus (Direct)    & 0.6016 & 0.2187 & 0.4869 & 0.3938 & 0.1600 & 0.5983 & 0.4650 & 0.4806 & 0.5405 & 0.2360 & 0.4181 \\
& Claude 4.5 Opus (DeepAgent) & 0.5929 & 0.1304 & 0.4107 & 0.3813 & 0.1067 & 0.6083 & 0.4850 & 0.4731 & 0.0000 & 0.2368 & 0.3425 \\
\bottomrule
\end{tabular}
\end{adjustbox}
\end{table*}

\section{Implications and Future Directions for Climate-Service Agents}
\label{sec:appendix_future_agents}

Modern agent systems offer multiple paradigms for strengthening service capability---tool use, subagents for specialized roles, subagents with fixed workflows (e.g., plan--extract--verify--compose), and reusable agent skills~\cite{anthropic_agent_skills_overview} that encapsulate reliable procedures.
In climate services, these components can be instantiated as concrete capabilities such as: code execution for quantitative checks and unit conversions; web/policy search for operational rules and advisories; retrieval over atmospheric reports and bulletins; and domain-database queries (e.g., reanalysis/forecast archives, hazard catalogs, thresholds and climatologies), all grounded by the service-product artifacts and metadata.

Our results suggest that the three core capabilities required for producing decision support deliverables each exhibits a distinct drawback that generic MLLMs and standardized scaffolds do not reliably resolve.
For actionable signal comprehension, errors concentrate in strict product interpretation and time localization; for decision-making handoff, failures are dominated by operationalization gaps in trigger/time clarity, feasibility/constraints, and uncertainty-to-action mapping; and for decision analysis \& planning, outputs become less stable and more topic-dependent for dynamic hazards, reflecting difficulty maintaining evidence-grounded, uncertainty-calibrated long-horizon reasoning.
These findings motivate capability-aligned agent components: product-aware comprehension skills and validators for time/region/risk slots; handoff-oriented workflows with constraint checkers, trigger generators, and uncertainty-conditioned branching templates; and planning subagents that retrieve domain context (reports, databases, policies) while enforcing non-overclaiming evidence rules.
Designing climate-service agents as a composition of such specialized components offers a concrete path to reliably produce decision support deliverables from operational S2S products.

%%%%%%%%%%%%%%%%%%%%%%%%%%%%%%%%%%%%%%%%%%%%%%%%%%%%%%%%%%%%%%%%%%%%%%%%%%%%%%%%%%%%%%%%%%%%%%%%
%%%%%%%%%%%%%%%%%%%%%%%%%%%%%%%%%%%%%%%%%%%%%%%%%%%%%%%%%%%%%%%%%%%%%%%%%%%%%%%%%%%%%%%%%%%%%%%%

\section{Conclusion}

We introduced \name, a multimodal benchmark for last-mile S2S climate services curated from an operational climate-service system, designed to evaluate whether MLLMs and agents can produce decision support deliverables from existing service products under uncertainty.
Across \nproducts products and \ncases cases, \name instantiates three service levels with schema-constrained outputs, yielding around \ntasks tasks and 1{,}000+ evaluation items.

% Benchmarking state-of-the-art MLLMs and a standardized agentic workflow shows that reliable service readiness remains challenging and strongly product-dependent.
% While models are often evidence-grounded, decision-facing deliverables frequently fail on operationalization---especially trigger/time clarity, feasibility/constraints, and uncertainty-to-action mapping---and generic scaffolding is not a consistent fix.
% We hope \name will accelerate the development of dedicated climate-service agents with evaluation-aligned guardrails and service-specific tooling.

Benchmarking state-of-the-art MLLMs and a standardized agentic workflow shows that reliable S2S service readiness remains challenging and strongly product-dependent. While models are often capable of evidence-grounded interpretation, they continue to struggle with S2S service plot understanding and reasoning-most notably actionable signal comprehension, operationalization for decision-making handoff (trigger/time clarity, feasibility/constraints, and uncertainty-to-action mapping), and stable evidence-grounded planning under dynamic hazards. These failures persist even with generic scaffolding, suggesting that progress will require service-specific training and tooling rather than prompting alone. We hope \name will accelerate the development of dedicated climate-service agents with evaluation-aligned guardrails and service-aware interfaces that directly target these bottlenecks.

\section{Acknowledgments}
We thank Xinyao Feng$^{1}$, Lu Tang$^{1}$, Yurong Song$^{2}$ and Lun Dai$^{2}$ for contributing a small set of expert-designed questions.

% 冯心瑶
% 汤璐
% 宋玉荣
% 代伦
$^1$Beijing Normal University \\
$^2$Nanjing University of Information Science and Technology \\

%%%%%%%%%%%%%%%%%%%%%%%%%%%%%%%%%%%%%%%%%%%%%%%%%%%%%%%%%%%%%%%%%%%%%%%%%%%%%%%
%%%%%%%%%%%%%%%%%%%%%%%%%%%%%%%%%%%%%%%%%%%%%%%%%%%%%%%%%%%%%%%%%%%%%%%%%%%%%%%
% Impact Statement
%%%%%%%%%%%%%%%%%%%%%%%%%%%%%%%%%%%%%%%%%%%%%%%%%%%%%%%%%%%%%%%%%%%%%%%%%%%%%%%
%%%%%%%%%%%%%%%%%%%%%%%%%%%%%%%%%%%%%%%%%%%%%%%%%%%%%%%%%%%%%%%%%%%%%%%%%%%%%%%

%\section*{Impact Statement}

%This paper advances machine learning evaluation by introducing \name, a benchmark for assessing multimodal decision-support generation for last-mile S2S climate services. Potential societal benefits include more transparent measurement of model reliability for climate resilience applications. Potential risks include over-reliance on model-generated analyses in high-stakes settings, dataset-driven overfitting, and bias or coverage gaps in curated cases. We recommend using \name strictly for evaluation and research, with human oversight and domain-specific validation for any real-world deployment.

%%%%%%%%%%%%%%%%%%%%%%%%%%%%%%%%%%%%%%%%%%%%%%%%%%%%%%%%%%%%%%%%%%%%%%%%%%%%%%%
%%%%%%%%%%%%%%%%%%%%%%%%%%%%%%%%%%%%%%%%%%%%%%%%%%%%%%%%%%%%%%%%%%%%%%%%%%%%%%%
% Bib
%%%%%%%%%%%%%%%%%%%%%%%%%%%%%%%%%%%%%%%%%%%%%%%%%%%%%%%%%%%%%%%%%%%%%%%%%%%%%%%
%%%%%%%%%%%%%%%%%%%%%%%%%%%%%%%%%%%%%%%%%%%%%%%%%%%%%%%%%%%%%%%%%%%%%%%%%%%%%%%

% \nocite{langley00}

\bibliography{main}
\bibliographystyle{unsrt}

%%%%%%%%%%%%%%%%%%%%%%%%%%%%%%%%%%%%%%%%%%%%%%%%%%%%%%%%%%%%%%%%%%%%%%%%%%%%%%%
%%%%%%%%%%%%%%%%%%%%%%%%%%%%%%%%%%%%%%%%%%%%%%%%%%%%%%%%%%%%%%%%%%%%%%%%%%%%%%%
% APPENDIX
%%%%%%%%%%%%%%%%%%%%%%%%%%%%%%%%%%%%%%%%%%%%%%%%%%%%%%%%%%%%%%%%%%%%%%%%%%%%%%%
%%%%%%%%%%%%%%%%%%%%%%%%%%%%%%%%%%%%%%%%%%%%%%%%%%%%%%%%%%%%%%%%%%%%%%%%%%%%%%%

\newpage
\appendix
% \onecolumn
% \section{You can have an appendix here.}

% You can have as much text here as you want. The main body must be at most $8$
% pages long. For the final version, one more page can be added. If you want, you
% can use an appendix like this one.

% The $\mathtt{\backslash onecolumn}$ command above can be kept in place if you
% prefer a one-column appendix, or can be removed if you prefer a two-column
% appendix.  Apart from this possible change, the style (font size, spacing,
% margins, page numbering, etc.) should be kept the same as the main body.

% In preamble:
% \usepackage{booktabs}
% \usepackage{array}

% ========== ========== ========== ========== 
{
\centering

\textbf{\Large Appendix}
\\
}

\section{Background: From Forecasts to Operational Service Products}
\label{sec:appendix_forecast_to_products}

\textbf{From climate forecasts to service products.}
S2S sits in the critical window between weather and seasonal outlooks (roughly 2 weeks to 2 months), where decisions often require both actionable lead time and explicit uncertainty quantification~\cite{yang2026last}. In today’s ecosystem, \textbf{S2S Prediction Providers} produce climate forecasts (often large-ensemble, probabilistic model outputs), while \textbf{S2S Service Providers} translate these forecasts into S2S service products tailored to sectoral decision contexts~\cite{yang2026last, hewitt2021climate, manrique2023subseasonal, terrado2022towards}. Concretely, service providers map forecast variables into sector-specific indices and impact models (e.g., fire danger indices, SPI drought indicators, navigability measures), and deliver products such as outlook/risk/action maps annotated with probabilities, decision thresholds, and human-interpretable confidence qualifiers~\cite{yang2026last}. The review further formalizes this translation as a progressive pipeline from deterministic prediction maps (with hindcast-based confidence), to tailor-made prediction maps (with historical confidence), and ultimately to action-informative maps that expose ensemble hit probabilities and threshold choices for decision support~\cite{yang2026last}. Despite these advances, the gap persists because the remaining burden---interpreting uncertain products, contextualizing them with local constraints and policies, and converting them into concrete actions---still largely falls on end users, making S2S an ideal and high-impact domain to evaluate and develop AI agents for decision-oriented climate services~\cite{yang2026last}. 
% More broadly, climate service providers face a persistent challenge in communicating climate-related information in ways that are understandable and actionable for both expert and non-expert audiences (Haase et al., 2000; Stephens et al., 2012; Ash et al., 2014; Nocke, 2014). Moreover, improved user uptake depends not only on higher S2S forecast skill (e.g., via better representations of physical processes and teleconnections; Merryfield, 2020), but also on downstream innovations in product design, post-processing, and operational workflows that translate model outputs into actionable climate information (Buontempo et al., 2014; Hewitt and Stone, 2021).

% Why is this interesting and important?
% \textbf{Why is this interesting and important?}
\textbf{From service products to actionable decisions.}
While S2S service products substantially lower the barrier from raw climate forecasts to sectoral interpretation, they still stop short of the final step that end users care about: translating uncertain, time-localized signals into concrete, context-aware decision support deliverables under real constraints (e.g., policies, resources, risk tolerances, and operating procedures). In practice, this last-mile step requires jointly reasoning over multimodal products (maps/plots/reports), uncertainty and confidence qualifiers, and domain-specific operational rules---capabilities that generic multimodal QA benchmarks do not capture. As a result, to build an S2S Service Agent that can reliably operate on top of existing service products, we first need an evaluation suite that is service-grounded (curated from operational climate-service products), decision-facing (measures usefulness and communication), and deliverable-oriented (tests whether outputs match operational artifacts).

\textbf{Evaluating service readiness.}
A central design choice in \name is to evaluate not only what a model answers, but also whether it can produce outputs in forms that resemble operational deliverables.
Accordingly, all tasks in \name use schema-constrained structured outputs to mirror machine-consumable handoffs in real workflows, while varying the granularity of the deliverable.
Specifically, \name includes \textbf{(\underline{i})} Short-slot Structured Completion (SSC) tasks that require short, triggerable fields (e.g., booleans, enumerations, numbers, or region lists) for early warning and signal extraction, and \textbf{(\underline{ii})} Structured Report Generation (SRG) tasks where longer decision-facing content is still organized into predefined sections/slots (often including multiple candidate plans/options) and graded with a rubric (Appendix~\ref{sec:oed_rubric}).
This unified structured interface enables fine-grained diagnosis of operational failure modes (e.g., multimodal misread vs.\ time-window errors vs.\ uncertainty misuse vs.\ schema noncompliance) under both direct prompting and standardized agentic workflows.

Using \name, we benchmark state-of-the-art MLLMs and agents, and analyze performance across service products and service levels, highlighting persistent challenges in S2S service plot understanding and reasoning such as actionable signal comprehension, operationalization for decision-making handoff (trigger/time clarity, feasibility/constraints, uncertainty-to-action mapping), and stable evidence-grounded planning for dynamic hazards.
Together, these analyses provide actionable guidance for building future climate-service agents and for designing training or tooling interventions that target the observed bottlenecks.

\section{Benchmark Size and Item-Instance Accounting}
\label{sec:appendix_benchmark_size}

This appendix reports the detailed size breakdown of \name at the granularity of product and service level. 
Table~\ref{tab:item-instances} summarizes the number of curated cases per product and the number of evaluation items instantiated at Levels I--III. 
We report total item instances computed as $\text{Cases}\times(\text{L1}+\text{L2}+\text{L3})$, which corresponds to the total number of checkable evaluation slots produced by the benchmark under our three-level formulation.

\begin{table}[h]
\centering
\caption{Total item instances computed as $\text{Cases}\times(\text{L1}+\text{L2}+\text{L3})$. L1, L2, L3 denote Level 1, Level 2, and Level 3, respectively; NDVI denotes for Normalized Difference Vegetation Index.}
\label{tab:item-instances}
\renewcommand{\arraystretch}{1.15}
\begin{adjustbox}{max width=0.7\textwidth}
\begin{tabular}{>{\raggedright\arraybackslash}p{5.6cm} r r r r r r}
\toprule
\textbf{Product} & \textbf{Cases} & \textbf{L1} & \textbf{L2} & \textbf{L3} & \textbf{Items per case} & \textbf{Total item instances} \\
\midrule
Drought & 21 & 4 & 3 & 2 & 9  & 189 \\
Crop Yield & 12 & 3 & 3 & 3 & 9  & 108 \\
Wind Energy & 16 & 1 & 1 & 1 & 3  & 48 \\
Atmospheric River & 19 & 3 & 3 & 3 & 9  & 171 \\
Solar Energy & 14 & 3 & 2 & 2 & 7  & 98 \\
Extreme Precipitation & 14 & 1 & 1 & 1 & 3  & 42 \\
Extratropical Cyclone & 10 & 4 & 4 & 1 & 9  & 90 \\
NDVI & 10 & 3 & 2 & 2 & 7  & 70 \\
Temperature & 20 & 8 & 1 & 1 & 10 & 200 \\
Tropical Cyclone & 25 & 2 & 1 & 1 & 4  & 100 \\
\midrule
\textbf{Total} & \textbf{161} & \textbf{32} & \textbf{21} & \textbf{17} &  & \textbf{1116} \\
\bottomrule
\end{tabular}
\end{adjustbox}
\end{table}

% ========== ========== ========== ========== 

\section{Structured Output Settings}
\label{sec:structured_output_settings}

\textbf{Unified structured-output interface for direct prompting.}
Although our direct prompting setting does not use agentic scaffolding, we still wrap each MLLM call with a unified structured-output interface to standardize schema-constrained deliverables across providers.
Specifically, we use the LangChain/LangGraph \texttt{with\_structured\_output()} interface to request structured outputs from MLLMs in a consistent manner~\cite{langchain_models_structured_output}.
For each task, we implement the target structured deliverable as a \textbf{Pydantic model}, and pass this model to \texttt{with\_structured\_output()}.

\textbf{Why Pydantic.}
Pydantic models provide a rich feature set for structured generation, including field validation, descriptions, and nested structures that closely match our benchmark schemas~\cite{pydantic_models_basic_usage}.
This allows us to enforce type constraints and required fields uniformly across tasks and models, and simplifies downstream parsing and evaluation.

\textbf{Gemini 3 Pro exception.}
We found that \texttt{with\_structured\_output()} is not compatible with our Gemini 3 Pro setup.
Therefore, for Gemini 3 Pro we use the native Google Gemini API for structured generation~\cite{google_ai_gemini_api_docs}, while keeping the same benchmark schemas and evaluation logic.

% ========== ========== ========== ========== 

% \section{Agreement with Expert Human Evaluation}
% \label{sec:human_agreement}

% TODO: We will report agreement between LLM-as-Judge and expert human graders.
% This section will include:
% (i) annotation protocol and sampled subset,
% (ii) agreement metrics (e.g., Spearman/Pearson correlation, Kendall tau, or weighted kappa where applicable),
% (iii) qualitative disagreement analysis and representative examples.

\section{Catalog of Curated Service Products and Domain Coverage}
\label{sec:appendix_product_catalog}

\textbf{Product-to-domain coverage.}
Table~\ref{tab:domain-mapping} summarizes the six-domain coverage of the curated operational products in \name.
A single product may support multiple decision contexts and is therefore associated with multiple domains when applicable.

\textbf{Curated service products.}
We provide brief descriptions of each product to clarify the operational intent and the types of signals exposed in the multimodal artifacts.

\begin{itemize}
    \item \textbf{Drought Outlook.}
    The drought forecasting product, based on the Standardized Precipitation Index (SPI), provides outlook maps and drought risk assessments derived from large-ensemble subseasonal predictions.
    It enables water managers to identify high-confidence drought signals and issue early warnings for mitigation.

    \item \textbf{NDVI Anomaly Outlook.}
    The NDVI-based agricultural product provides next-month risk levels (low/medium/high) based on NDVI anomalies.
    It supports agricultural management by highlighting areas likely to experience vegetation stress relative to climatology.

    \item \textbf{Crop Yield Outlook.}
    This product offers spatially explicit crop yield predictions, supporting agricultural planning, food security assessment, and market analysis.

    \item \textbf{Potential Solar Energy Outlook.}
    The solar energy product generates monthly outlook maps of solar resource distributions and probabilities of low-light events.
    It supports renewable planning and risk management for weather-sensitive generation.

    \item \textbf{Potential Wind Energy Outlook.}
    The wind energy product provides probabilistic forecasts for wind resources, facilitating renewable energy planning and operational scheduling.

    \item \textbf{Tropical Cyclone Outlook.}
    This product visualizes tropical cyclone (TC) paths with auxiliary metrics such as sea surface temperature and intensity category.
    It supports disaster preparedness, risk assessment, and operational monitoring.

    \item \textbf{Extratropical Cyclone Frequency Outlook.}
    This product summarizes the pathways, intensity, and frequency of extratropical cyclones on a monthly basis using detection algorithms.
    It supports climate monitoring and preparedness planning for mid-latitude hazards.

    \item \textbf{Extreme Precipitation Frequency.}
    This product provides monthly forecasts of the cumulative frequency of extreme precipitation, where extremes are defined relative to model climatology and counted as days exceeding the threshold.

    \item \textbf{Atmospheric River Outlook.}
    The atmospheric river (AR) product detects AR pathways and intensity from hindcast and forecast data, producing weekly/monthly activity and anomaly maps.
    It also includes basin-scale summaries (e.g., moisture contribution time series) to support flood preparedness and water-resource management.

    \item \textbf{Temperature Monitoring.}
    This product provides temperature outlooks including spatial anomaly maps and ensemble-based time-series summaries, highlighting trends and uncertainty for heat-risk monitoring and energy/agriculture applications.
\end{itemize}

\section{Service Task Level Definitions}
\label{sec:appendix_levels}

\textbf{Definition schema.}
Each service level is specified by \textbf{Goal}, \textbf{Output format}, and \textbf{Evaluated items}.
Levels~II--III additionally specify \textbf{Required evidence} to enforce product-grounded service reasoning under uncertainty.

\begin{itemize}[leftmargin=1em, topsep=3pt]
    \item \textbf{Level I: Signal comprehension --- Early Warning \& Mitigation (EWM).}
    \textbf{Goal}: extract decision-relevant signals from the service product in an actionable and machine-consumable form (e.g., impacted regions, risk tiers, confidence cues) with correct time localization.
    \textbf{Output format}: strictly schema-constrained JSON (SSC); free-form justification is disallowed.
    \textbf{Evaluated items}: schema compliance and slot-level correctness/completeness.

    \item \textbf{Level II: Action Handoff --- decision-making handoff (DMH).}
    \textbf{Goal}: produce an operationally usable, decision-making handoff that translates product signals into executable response guidance under uncertainty, including explicit scenario branches (e.g., high/medium/low confidence; best/base/worst) and contingency actions with clear triggers, timing, and escalation/de-escalation rules.
    \textbf{Required evidence}: the output must \textbf{(\underline{i})} remain faithful to the provided service products (signals, valid-time windows, affected regions, risk tiers, confidence cues), and \textbf{(\underline{ii})} explicitly map uncertainty/confidence to action intensity and branching (low-regret vs.\ committed actions), without overclaiming or contradicting the product.
    \textbf{Output format}: schema-constrained report-style handoff (SRG) with predefined fields/slots; when applicable, multiple scenario/option slots are required.
    \textbf{Evaluated items}: rubric-based grading emphasizing trigger/time clarity, feasibility/constraints, uncertainty handling, actionability, and evidence grounding (Appendix~\ref{sec:oed_rubric}).

    \item \textbf{Level III: Strategic Assessment --- Decision Analysis \& Planning (DAP).}
    \textbf{Goal}: provide a reasoning-based strategic assessment that explains and interprets the spatial--temporal patterns present in the service products, identifies key anomalous regions/signals and their potential implications, and proposes planning-oriented considerations (e.g., monitoring priorities, preparedness posture, longer-horizon risk management) consistent with the provided evidence.
    \textbf{Required evidence}: analyses must be grounded in the provided service-product artifacts; any explanatory hypotheses (e.g., large-scale drivers or mechanisms) must be stated as plausible and non-overclaiming interpretations rather than unverifiable facts, and must not override or contradict the product signals.
    \textbf{Output format}: schema-constrained report-style deliverable (SRG) with predefined analysis/planning fields/slots (e.g., key signals, implications, hypotheses, planning notes, uncertainty/limits).
    \textbf{Evaluated items}: rubric-based grading emphasizing evidence grounding, uncertainty calibration, clarity of implications and planning guidance, and operational feasibility (Appendix~\ref{sec:oed_rubric}).
\end{itemize}

\textbf{Illustrative examples.}
We provide representative task prompts at each service level to illustrate the progression from signal comprehension to operational handoffs and strategic assessment.

\begin{itemize}[leftmargin=1em, topsep=2pt]

  \item \textbf{Level I (EWM; Signal Comprehension).}
  Tasks focus on triggerable signal extraction from multimodal service products under strict structure.
  Examples include:
  \textbf{(\underline{i})} \textbf{Atmospheric River (AR)}: identify high-risk areas that simultaneously satisfy a frequency threshold (e.g., $\ge$6 days) and high ensemble agreement (e.g., $>50\%$), and output the location names, probability range, and risk level, or mark \texttt{Not triggered} if none;
  \textbf{(\underline{ii})} \textbf{Drought}: extract latitude--longitude ranges of high-risk areas and characterize high-confidence regions using hatched overlays;
  \textbf{(\underline{iii})} \textbf{Yield}: identify high/medium/low-yield spatial regions by color coding and report their latitude--longitude ranges;
  \textbf{(\underline{iv})} \textbf{Solar}: extract latitude--longitude ranges for abundant solar resources and for solar drought risk regions.

  \item \textbf{Level II (DMH; Action Handoff).}
  Tasks require producing decision-making handoff grounded in the recognized risk and confidence signals, with explicit actions and contingency branches.
  Examples include:
  \textbf{(\underline{ii})} \textbf{Drought}: propose emergency response measures conditioned on risk tiers and confidence levels (e.g., high-risk/high-confidence vs.\ medium-to-high-risk/low-confidence), and specify contingency actions for low-confidence areas under drought beyond expectations and sudden precipitation surges;
  \textbf{(\underline{ii})} \textbf{Yield}: integrate predicted yield timing and spatial patterns of high/low yield to produce actionable field management recommendations that safeguard yield stability;
  \textbf{(\underline{iii})} \textbf{Solar}: propose targeted response measures for high-confidence solar drought regions with existing photovoltaic installations, and report the corresponding latitude--longitude coordinates together with the response strategies;
  \textbf{(\underline{iv})} \textbf{AR}: integrate an AR action-informative map with geographic and social factors to deliver a management plan, and determine whether forecasted AR risk represents a Flood Risk (defense) or a Water Resource Opportunity (collection) for key regions.

  \item \textbf{Level III (DAP; Strategic Assessment).}
  Tasks emphasize in-depth analysis and planning beyond immediate handoffs, such as diagnosing anomalies, assessing broader suitability, and recommending longer-horizon strategies.
  Examples include:
  \textbf{(\underline{i})} \textbf{Drought}: assess the severity of the current drought situation and provide recommendations for long-term drought disaster prevention planning;
  \textbf{(\underline{ii})} \textbf{Yield}: propose a comprehensive and operational management framework under low-yield conditions, together with policy-oriented recommendations;
  \textbf{(\underline{iii})} \textbf{Solar}: analyze current solar availability and solar drought risks in regions with concentrated photovoltaic power stations (in the context of long-term climatology) and assess suitability for continued development;
  \textbf{(\underline{iv})} \textbf{AR}: review global AR frequency and probability maps, identify the most anomalous regions relative to climatology, and infer plausible large-scale drivers (e.g., ENSO phase, jet stream shifts) using meteorological reasoning.

\end{itemize}

\begin{table*}[t]
\centering
\small
\begin{tabular}{p{0.55\linewidth} p{0.40\linewidth}}
\toprule
\textbf{Product} & \textbf{Domain label(s)} \\
\midrule
Drought & Agriculture; Disasters; Finance \\
Tropical Cyclone & Disasters; Shipping; Finance; Energy \\
Extratropical Cyclone & Disasters; Shipping; Energy \\
Atmospheric River & Disasters; Energy; Agriculture \\
Extreme Precipitation & Disasters; Agriculture; Finance \\
% Thunderstorm Activity & Disasters; Energy; Shipping \\
Temperature & Health; Energy; Agriculture \\
Normalized Difference Vegetation Index (NDVI) & Agriculture; Finance \\
Crop Yield & Agriculture; Finance \\
% Marine Fisheries & Shipping; Agriculture; Finance \\
Solar Energy & Energy \\
Wind Energy & Energy \\
% Hydropower Potential & Energy; Disasters \\
% Thermal Power Load Forecasting & Energy; Health \\
\bottomrule
\end{tabular}
\caption{Product-to-domain coverage in \name. A single operational product may support multiple decision contexts and is therefore associated with multiple domains when applicable.}
\label{tab:domain-mapping}
\end{table*}

\section{Evaluation Details}
\label{sec:appendix_evaluation_details}

\subsection{SSC Evaluation: Schema Validity and Field Scoring}
\label{sec:ssc_eval}

\textbf{Schema compliance.}
For SSC tasks, a prediction must parse as valid JSON and conform to the required schema (required keys present; types match; enumerations valid).
If the output fails schema compliance, we assign zero to the corresponding missing/invalid fields and treat the schema-compliance indicator as failed.

\textbf{Field-level scoring by type.}
SSC tasks contain three common field types:
\textbf{(i) booleans}, \textbf{(ii) numbers}, and \textbf{(iii) strings}.
We score each required field and then aggregate to obtain task-level SSC scores.

\begin{itemize}[topsep=2pt,leftmargin=1.2em]
    \item \textbf{Boolean fields.} Exact match with the reference (\texttt{true}/\texttt{false}).
    \item \textbf{Numeric fields.} Correct if the absolute relative error is within 5\%:
    $
    \frac{| \hat{x} - x |}{\max(|x|,\epsilon)} \le 0.05
    $
    where $x$ is the reference value, $\hat{x}$ is the model prediction, and $\epsilon$ is a small constant to avoid division by zero.
    \item \textbf{String fields.}
    All string-valued fields are scored using \textbf{LLM-as-Judge} with a fixed judging protocol.
    This is primarily to handle format variation and semantic equivalence that are common in operational service outputs (e.g., synonymous risk labels, alternate naming/abbreviations for regions, mixed granularity, and multi-region descriptions).
    For categorical strings in a closed set (e.g., \texttt{risk\_category} $\in$ \{low, medium, high\}), the judge performs normalization and returns an equivalence verdict that is functionally equivalent to exact match.
    For region/area strings (e.g., \texttt{high\_confidence\_region}, \texttt{high\_risk\_area}), the judge normalizes the predicted and reference regions to a comparable set of geographic units and returns an overlap-based coverage score (defined below).
\end{itemize}

\textbf{Coverage-based matching for region/area strings.}
For region/area fields, exact string match is brittle due to variations in naming, granularity, and multi-region descriptions.
We therefore use the judge to normalize predicted and reference regions to a comparable set of geographic units and compute an overlap-based coverage score.
Concretely, let $R$ be the reference region set and $\hat{R}$ be the predicted region set after normalization; we define the field score as reference coverage
$
\mathrm{Cov}(\hat{R},R)=\frac{|\hat{R}\cap R|}{|R|}.
$
This design emphasizes whether the model covers the expert-identified impacted area, aligning with last-mile use cases where missing affected regions can be deployment-critical.
(We use the same judge prompt and normalization protocol for all models.)

\textbf{Task-level aggregation.}
SSC task scores are computed by averaging required-field scores within a task.
We report macro-averages across tasks, and also break down results by product, domain, and service level.

\subsection{SRG Evaluation: Rubric-Based Services-Oriented Grading}
\label{sec:srg_eval}

\textbf{Rubric and judge model.}
SRG tasks are graded using the service-oriented rubric in Appendix~\ref{sec:oed_rubric}, which scores each output on six dimensions (0--5 each) and reports a total (0--30) and average (0--5). 
% For Level~III (DMH), we include an optional field \texttt{historical\_context\_addendum} that allows models to provide brief historical analogs or climatological background when helpful. This field is not required for compliance and is reported as an auxiliary diagnostic rather than part of the primary rubric score.
% For Level~III (DMH), models may optionally provide a \texttt{historical\_context\_addendum}. This addendum is not required and does not affect the primary 6-dimension rubric score. We optionally analyze its usefulness/faithfulness as a diagnostic to understand when background context helps or hurts decision-making handoff.
We use GPT-5.2 as the judge with a fixed evaluation prompt and deterministic decoding (temperature set to 0) to output dimension-wise scores (and optional brief rationales).
All models are graded under the same rubric and judging protocol.

\textbf{Minimum Structure Requirement (MSR).}
To ensure SRG outputs are comparable and scorable across models, we append a Minimum Structure Requirement (Appendix~\ref{sec:oed_rubric}) to every SRG prompt.
MSR constrains output structure only (minimum required fields/content categories) and does not provide additional evidence beyond the given service artifacts.
This prevents degenerate, underspecified responses and improves evaluation consistency, while preserving the core requirement: producing decision-making handoff grounded in the provided operational products.

\textbf{Scoring and aggregation.}
For each SRG task, the judge outputs a 0--5 score for each rubric dimension.
When a task includes multiple candidate plans/options, we grade each plan/option separately and average across plans/options to obtain the task-level SRG score.
We report dimension-wise averages to diagnose which aspects of service readiness (e.g., triggers/time clarity, uncertainty handling) are most limiting.

\section{SRG LLM-as-Judge Rubric}
\label{sec:oed_rubric}
% --- Preamble (add if you don't have them) ---

% -----------------------------
% Rubric Table (SAFE VERSION)
% -----------------------------
\subsection{Rubric Table: SRG services-oriented Rubric (6 Dimensions $\times$ 0--5)}

\noindent\textbf{Scoring:} Each dimension is an integer 0--5.\\
\textbf{Total} = sum (0--30), \textbf{Average} = Total/6 (0--5).

\begin{table}[ht]
\centering
\scriptsize
\renewcommand{\arraystretch}{1.25}

\resizebox{\textwidth}{!}{%
\begin{tabular}{p{3.0cm} p{4.2cm} p{3.2cm} p{3.2cm} p{3.0cm} p{2.8cm} p{3.2cm}}
\toprule
\textbf{Dimension (0--5)} &
\textbf{5 --- Excellent (service-ready)} &
\textbf{4 --- Strong} &
\textbf{3 --- Adequate} &
\textbf{2 --- Weak} &
\textbf{1 --- Minimal} &
\textbf{0 --- Not present / invalid} \\
\midrule

\textbf{1) Context Tailoring (Geo / Sector / Population / Asset)} &
$\ge$3 concrete tailored strategies across \textbf{(a) geography/environment}, \textbf{(b) sector/asset operations}, \textbf{(c) population/vulnerability}, with clear ``why it matters''. &
$\ge$2 tailored strategies covering at least 2 categories; rationale mostly clear. &
One concrete tailoring; partial coverage or weak rationale. &
Vague ``consider local context'' with little actionable tailoring. &
Generic mention only. &
Fully generic template. \\

\textbf{2) Actionability (Concrete Measures)} &
Structured plan with \textbf{$\ge$8 actionable measures} spanning multiple domains (monitoring, operations, protection, logistics, comms), clearly differentiated by scenario (risk/confidence/region). &
$\ge$6 actions; $\ge$3 domains; scenario differentiation mostly clear. &
$\ge$4 actions but generic and/or missing key domains. &
1--3 actions and/or mostly slogans. &
Almost no executable actions. &
Off-topic / no actions. \\

\textbf{3) Trigger \& Time/Horizon Clarity} &
Clear \textbf{activation--escalation--de-escalation} timing/horizon (hours--weeks--years as appropriate) \textbf{and $\ge$3 operational triggers} (thresholds/conditions). &
Clear timing + $\ge$2 triggers. &
Either timing or triggers is clear, not both. &
Vague timing; no usable triggers. &
Scattered mention; not operational. &
No timing/horizon or triggers. \\

\textbf{4) Evidence Grounding (Link to Inputs/Evidence)} &
Explicitly ties key actions to \textbf{provided inputs} (e.g., risk/confidence results, multimodal evidence), with consistent ``action $\leftrightarrow$ evidence'' alignment. &
Most actions grounded; minor gaps. &
Overall grounding stated but not consistently linked to actions. &
Only generic ``based on forecast''; actions not grounded. &
Nearly no justification. &
Contradictory/hallucinated rationale. \\

\textbf{5) Feasibility \& Constraints (Operational/Policy/Resource Limits)} &
Identifies $\ge$2 realistic constraints (capacity, resources, coordination, safety, regulations, cost/time-lag) and provides feasible alternatives/prioritization. &
Identifies $\ge$1 key constraint + mitigation strategy. &
Mentions constraints but generic handling. &
Token mention only. &
Unrealistic / ignores feasibility. &
Dangerous or infeasible recommendations. \\

\textbf{6) Uncertainty \& Confidence Handling (Robustness / Contingency)} &
Uses confidence to modulate actions (low-regret vs committed), provides \textbf{contingency branches} for plausible adverse surprises, and specifies escalation rules as confidence updates. &
Correct low-confidence strategy; branches mostly complete but escalation rules unclear. &
Mentions uncertainty/monitoring but branching incomplete. &
Only states uncertainty exists. &
Misuses confidence. &
No uncertainty handling. \\

\bottomrule
\end{tabular}%
}
\end{table}

% --------------------------------------------
% Universal “Minimum Structure Requirement”
% --------------------------------------------
\subsection{Universal ``Minimum Structure Requirement'' (Append to Any SRG Prompt)}

\noindent\textbf{Minimum structure requirement (must follow):}
\begin{enumerate}
  \item \textbf{Context (tailoring):} State at least \textbf{two} relevant contextual factors drawn from \textbf{geography/environment}, \textbf{sector/assets}, and/or \textbf{population/vulnerability}.
  \item \textbf{Actions (what to do):} Provide at least \textbf{six} concrete, executable measures, grouped into the following sections:
  \begin{itemize}
    \item \textbf{(\underline{a}) Monitoring \& early warning}
    \item \textbf{(\underline{b}) Operational/sector actions} (choose what fits: water/energy/agriculture/transport/health/others)
    \item \textbf{(\underline{c}) Protection \& resource/logistics} (staffing, supplies, mutual aid, safety)
    \item \textbf{(\underline{d}) Public communication \& coordination}
  \end{itemize}
  \item \textbf{Triggers \& time/horizon (when to do it):} Provide at least \textbf{two} explicit triggers (thresholds/conditions) and specify the relevant \textbf{time horizon} (e.g., next 72 hours, week 3--4, next month, 2--5 years), including \textbf{escalation and de-escalation} rules.
  \item \textbf{Confidence \& robustness:} Explain how \textbf{confidence} changes action intensity (e.g., low-regret vs committed actions) and include at least \textbf{one contingency branch} for plausible adverse surprises.
  \item \textbf{Feasibility/constraints:} Mention at least \textbf{one} feasibility constraint (capacity, resources, coordination, safety, regulations, cost/time-lag) and how the plan adapts.
\end{enumerate}

\end{document}